\pdfoutput=1

\documentclass[12pt]{article}

\usepackage[caption=false]{subfig}
\usepackage{sbc-template}

\usepackage{needspace}
\usepackage{graphicx,url}
\usepackage{booktabs, array}

\usepackage[utf8]{inputenc}  

\usepackage{amsmath}
\usepackage{amsfonts}

\title{Portrait Stylization: Artistic Style Transfer with \\Auxiliary Networks for Human Face Stylization}

\author{Thiago Ambiel\inst{1}}

\address{Instituto de Ciências Matemáticas e de Computação -- University of São Paulo (USP)\\
CEP -- 13566-590 -- São Carlos -- SP -- Brazil
\email{thiago.ambiel@usp.br}
}

\begin{document} 

\maketitle

\begin{abstract}
    Today's image style transfer methods have difficulty retaining
    humans face individual features after the whole stylizing
    process.
    This occurs because the features like face
    geometry and people's expressions are not captured by the
    general-purpose image classifiers like the VGG-19 pre-trained models.
    This paper proposes the use of embeddings from an
    auxiliary pre-trained face recognition model to encourage
    the algorithm to propagate human face features from the
    content image to the final stylized result.
\end{abstract}

\section{Introduction} \label{sec:introduction}

    The style transfer technique is currently one of the most studied areas in deep learning applied to digital art.
    It consists in reconstructing a given \emph{content image} -- the image
    that will seed the overall structure of the resulting image,
    e.g.\ an animal, a landscape or a human portrait -- with the style of a given \emph{style image}
    -- the image that will seed the texture of the resulting image,
    e.g.\ an oil painting or an abstract art -- producing the final stylized \emph{result image} output.
    
    This technique made possible new forms of digital art, like reproducing classical paintings with
    the content of modern days or creating amazing new effects that weren't previously possible.
    Despite that, actual state-of-the-art algorithms show limitations when applied to images with human faces.
    These methods yield face deformations in the final result image, which makes it difficult to recreate
    portrait paintings such as \emph{The Mona Lisa from Leonardo da Vinci} or
    \emph{Self Portrait from Vincent Van Gogh}.
    
    The main cause of this is that the human face geometry is not passed as an important
    content criterion, as the VGG-Network~\cite{vgg_net} can't capture so much relevant
    features about the human face.
    This problem can be solved by simply using a face recognition model like
    FaceNet~\cite{facenet} as an auxiliary model for content feature extraction.
    So, in that way, the human face geometry and other relevant facial features
    can be propagated to the final \emph{result image}.

\section{Background} \label{sec:backgrond}

    This problem can be formulated as Finding the optimal changes for a \emph{content image}
    that minimizes the difference between its texture and the texture from a \emph{style image}
    without losing the high-level information contained in it.
    These content and style differences can be calculated through the internal representations of
    a pre-trained deep convolutional neural network like the VGG-Network given that,
    following the paper \emph{A Neural Algorithm of Artistic Style}~\cite{gatys_artistic_style}, the
    higher layers from these networks can capture high-level information -- like textures and color
    palettes -- from its input images, at the same time that its lower layers can capture low-level
    information like the object's geometry and its colors.

    There are many ways of finding these optimal changes, but here the focus goes
    to the traditional optimization technique.
    This technique was first introduced by Gatys' 2015 paper and
    uses the quasi-Newton optimization algorithm \emph{Limited-memory BFGS}
    optimizer~\cite{lbfgs_optim} to solve the style transfer problem by optimizing
    the pixels of an \emph{initial image}, that can be a noise or the own \emph{content image},
    in the same way that the weights of a deep learning model are optimized, through the
    minimization of a criterion function (Eq.~\ref{eq:eq_total}).

    This criterion function is the weighted sum between a \emph{content loss} function
    (Eq.~\ref{eq:eq_content}) and a \emph{style loss} function (Eq.~\ref{eq:eq_style}),
    This means that the content and style weights can be controlled by $\alpha$ and $\beta$
    factors respectively, giving the user more control over the \emph{result image}.
    The function is defined as follows:

    \begin{equation}
        \mathcal{L}_{total}(\vec{c}, \vec{s}, \vec{x}) = \alpha \mathcal{L}_{content}(\vec{c}, \vec{x}) + \beta \mathcal{L}_{style}(\vec{s}, \vec{x}) \label{eq:eq_total}
    \end{equation}
    
    Where $\vec{c}$ is the \emph{content image}, $\vec{s}$ the \emph{style image} and
    $\vec{x}$ the \emph{result image}.
    
    \subsection{Content Loss} \label{subsec:content-loss}
        The content of a given image can be represented by the responses from convolutional filters
        of lower layers in a pre-trained VGG-Network that will be called here \emph{feature representations}.
        The responses from a given layer $l$ can be stored in a matrix $F^l \in \mathbb{R}^{N_l \times M_l}$
        where $N_l$ represents the number of filters contained in layer $l$,
        $M_l$ represents the number of pixels -- product between width and height -- of the filters
        contained in layer $l$,
        and $F^l_{ij}$ are the output activations from the $i^{th}$ filter at position $j$ from layer $l$.
        
        In that way, the \emph{content loss} function for a given layer $l$ can be defined as the
        mean squared error between the \emph{content image feature representations} $C^l$ and the
        \emph{result image feature representations} $X^l$:
        
        \begin{equation}
            \mathcal{L}_{content}(\vec{c}, \vec{x}, l) = \frac{1}{N} \sum_{i,j}^{N} (C^l_{ij} - X^l_{ij})^2 \label{eq:eq_content_layer}
        \end{equation}
        
        Then, being $L$ the total number of content layers and $W_{l}^{c}$ the relative weight
        of content layer $l$, the final \emph{content loss} function can be calculated by
        the weighted sum of all the content layer losses:
        
        \begin{equation}
            \mathcal{L}_{content}(\vec{c}, \vec{x}) = \sum_{l=0}^L W_{l}^{c} \times \mathcal{L}_{content}(\vec{c}, \vec{x}, l) \label{eq:eq_content}
        \end{equation}

    \subsection{Style Loss} \label{subsec:style-loss}
        Like the content \emph{feature representations}, the style of a given image is also represented
        by the responses from convolutional filters of a pre-trained VGG-Network.
        Nevertheless, it is actually represented by higher layers of the network and not directly,
        but through the correlations between the filter responses.
        These correlations are given by the Gram Matrix $G^l \in \mathbb{R}^{N_l \times N_l}$, where:
        
        \begin{equation}
            G^l_{ij} = \sum_{k} F^l_{ik} \times F^l_{jk} \label{eq:eq_gram_matrix}
        \end{equation}
        
        From that way, the \emph{style loss} function from a given layer $l$ can be defined as the mean
        squared error between the Gram Matrix of layer $l$ from the \emph{style image} $S^l$ and
        the Gram Matrix of layer $l$ from the \emph{result image} $A^l$:
        
        \begin{equation}
            \mathcal{L}_{style}(\vec{s}, \vec{x}, l) = \frac{1}{4 N_l^2 M_l^2} \sum_{i,j} (S^l_{ij} - A^l_{ij})^2 \label{eq:eq_style_layer}
        \end{equation}
        
        Finally, being $L$ the total number of style layers and $W_{l}^{s}$ the relative weight
        of style layer $l$, the final \emph{style loss} function can be calculated by the weighted sum of all
        the style layer losses:
        
        \begin{equation}
            \mathcal{L}_{style}(\vec{s}, \vec{x}) = \sum_{l=0}^{L} W_{l}^{s} \times \mathcal{L}_{style}(\vec{s}, \vec{x}, l) \label{eq:eq_style}
        \end{equation}

    \subsection{The Problem of High Resolution} \label{subsec:problem}

\begin{figure}[htp]
            \centering
            \subfloat[Content Image]{
                \includegraphics[width=1.8in]{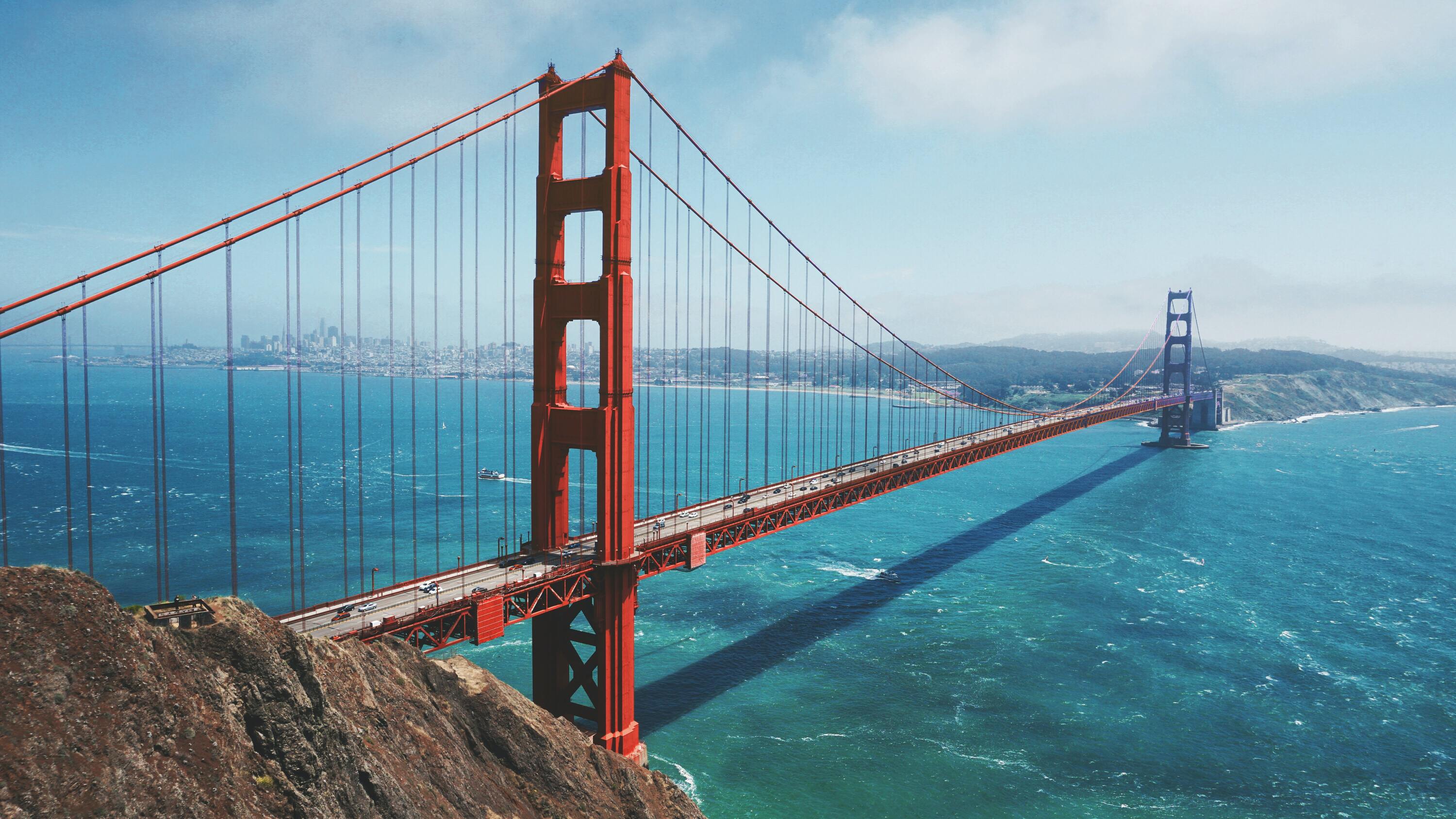}
                \label{fig:subfig-a}
            }
            \hspace{.1em}%
            \subfloat[Style 1]{
                \includegraphics[width=1.8in]{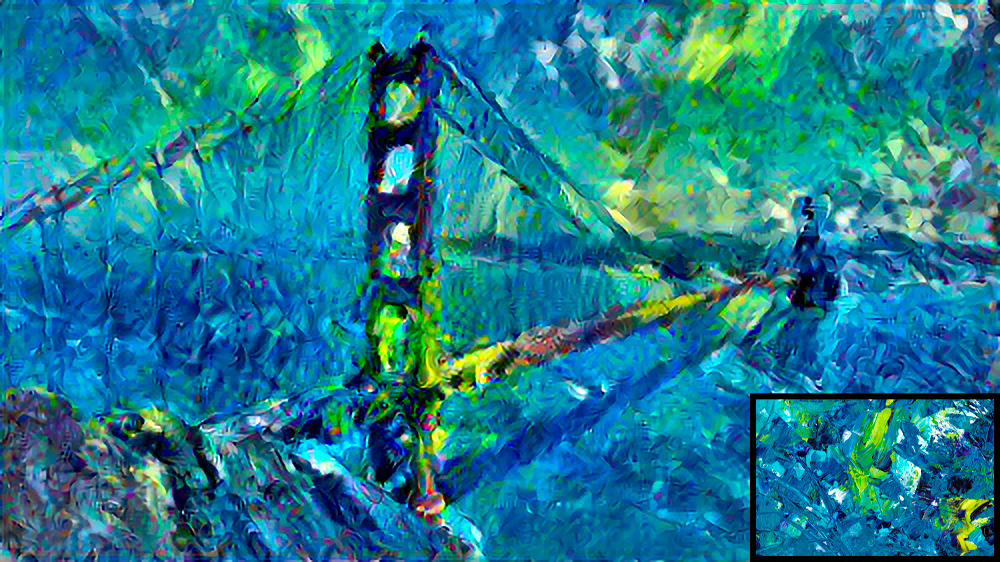}
                \label{fig:subfig-b}
            }
            
            \subfloat[Style 2]{
                \includegraphics[width=1.8in]{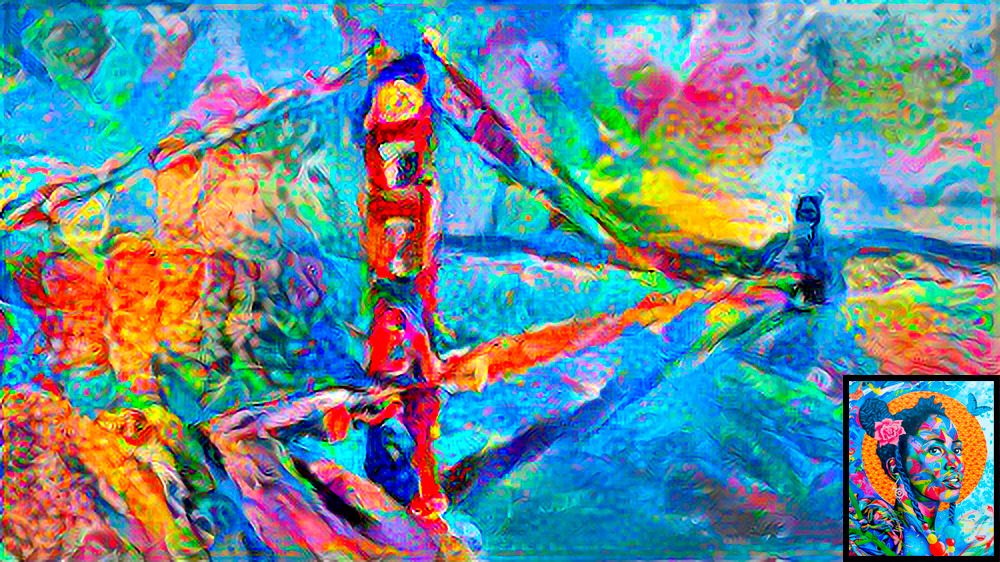}
                \label{fig:subfig-c}
            }
            \hspace{.1em}%
            \subfloat[Style 3]{
                \includegraphics[width=1.8in]{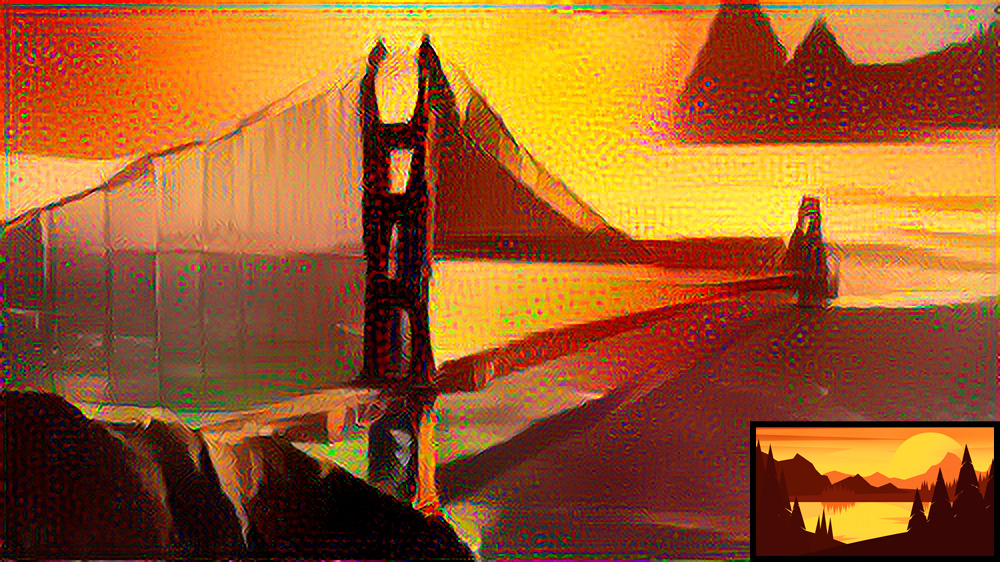}
                \label{fig:subfig-d}
            }
        
            \caption{
                An image from \emph{The Golden Gates Bridge} stylized with different
                \emph{style images} through the Gatys' 2015 method.
                All samples are generated at a \emph{500px} resolution, and each style
                used is minimized at the right-bottom edge of each sample.
            }
            \label{fig:gatys_results}
        \end{figure}

        Beautiful synthetic images can be generated by applying the Gatys'
        method to a given \emph{content image} and \emph{style image}, like
        the ones at Figure~\ref{fig:gatys_results}.
    
\begin{figure}[ht]
            \centering
            \subfloat[Style 1]{\includegraphics[width=1.8in]{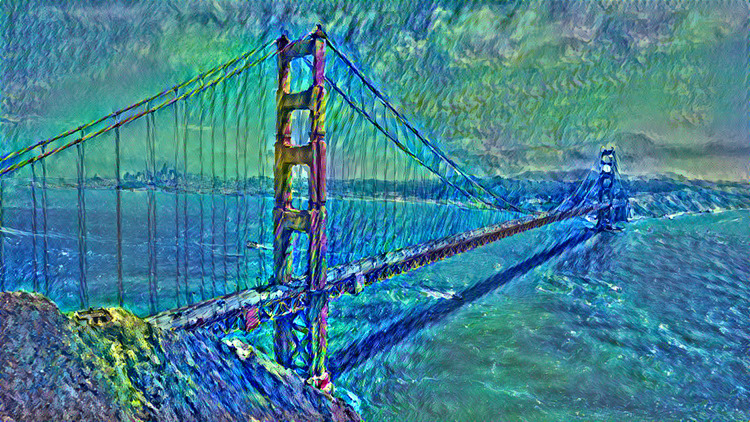}}%
            \hspace{.1em}%
            \subfloat[Style 2]{\includegraphics[width=1.8in]{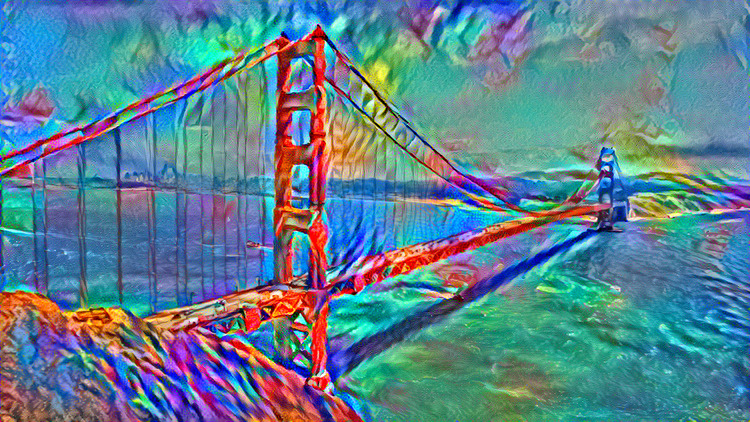}}%
            \hspace{.1em}%
            \subfloat[Style 3]{\includegraphics[width=1.8in]{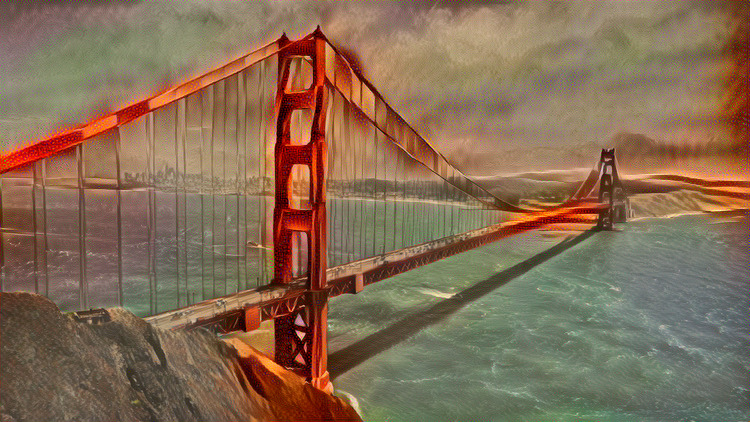}}
        
            \caption{
                An image from \emph{The Golden Gates Bridge} stylized with the respective
                \emph{style images} from Figure~\ref{fig:gatys_results} in high resolution (1080px).
                These samples show the limitations of the method when applied to
                high-resolution images.
            }
            \label{fig:gatys_fail_results}
        \end{figure}

        Nevertheless, this method has limitations when applied to high
        resolution images (e.g.\ 1080px or 1440px), resulting in only some
        color changes and almost no structural change (Figure~\ref{fig:gatys_fail_results}).
        This phenomenon occurs because the receptive fields of convolutional neural networks have
        fixed sizes, and in higher-resolution images its becomes relatively small,
        what makes the VGG-Network to pay attention only to small structures of the image
        during the stylization process, ignoring the overall image structure.
    
        Following the paper \emph{Controlling Perceptual Factors in Neural Style Transfer}
        ~\cite{gatys_hq_style_transfer}, the optimal resolution size of input images when using
        the VGG-Network for style transfer is around 500px, where bigger structural changes
        occurs while image content is well-preserved, like in Figure~\ref{fig:gatys_results}(c).

\section{Improving the Quality of the Generations} \label{sec:improving-the-results-quality}

    The Gatys' 2016 paper also proposes that the quality of generations at
    high resolution can be improved through the \emph{coarse-to-fine} technique.
    It consists of dividing the stylizing process into two stages, where given a
    high-resolution \emph{content} and \emph{style} images $\vec{c}$ and
    $\vec{s}$ with $N$ number of pixels, is defined by the following processes:

    In the first stage, the images are downsampled by a $K$ factor, such that
    $N/K$ corresponds to the resolution where the stylization will be performed,
    e.g.\ $500^2$px for VGG-Network.
    After this process, the stylization is performed with the downsampled
    images, generating a low-resolution \emph{result image} $\vec{z}$.

\begin{figure}[htp]
        \centering
        \subfloat[Low resolution (500px)]{
            \includegraphics[width=2.1in]{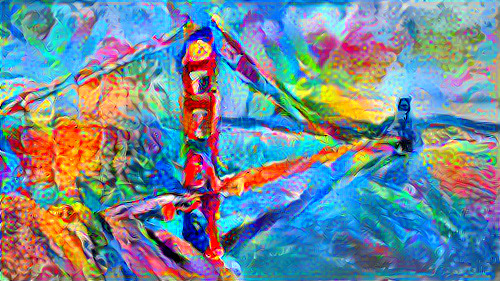}
        }
        \hspace{.1em}%
        \subfloat[Upsampled (1080px)]{
            \includegraphics[width=2.1in]{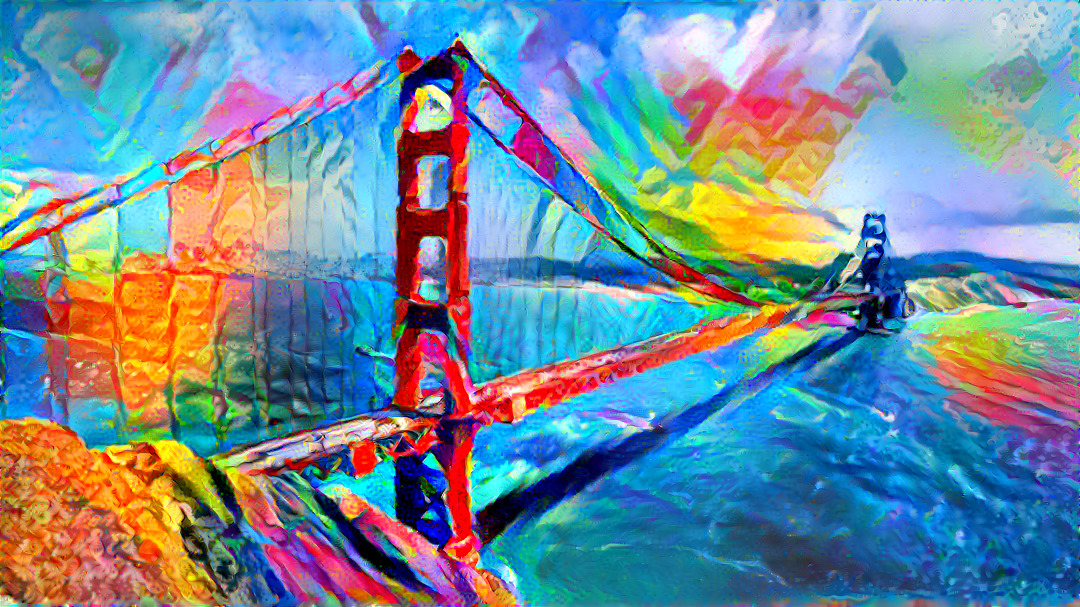}
        }
    
        \caption{
            An example of the \emph{coarse-to-fine} process applied
            to \emph{The Golden Gates Brigde} stylized image from
            Figure~\ref{fig:gatys_results}(c).
        }
        \label{fig:gatys_upsampling}
    \end{figure}

    Now, at the second stage, the generated image $\vec{z}$ is upsampled to
    $N$ pixels and then used as \emph{initial image} for the stylization process
    of the original input images $\vec{c}$ and $\vec{s}$, generating the final
    \emph{result image} $\vec{x}$.
    This process causes the algorithm to fill in the high-resolution information
    without losing the overall style structure of the $\vec{z}$ image (Figure~\ref{fig:gatys_upsampling}(b)).

    The method proposed in this paper is based on Crowson's Neural Style
    Transfer implementation~\cite{crowson_style_transfer}.
    Various relevant changes were made to the stylizing process to
    improve the quality of the generated images, but here only some
    of those will be discussed.

    First, the \emph{coarse-to-fine} technique is divided into $s$ stages,
    rather than one stage for low-resolution stylization and another to
    high-resolution one.
    The stylization process is started at an \emph{initial resolution} $r_{i}$
    and then is applied to progressively larger scales, each greater by a factor
    $k = \sqrt 2$ until it reaches a \emph{final resolution} $r_{f}$.
    In this way, the resolution $R(s)$ at a given stage $s$ is defined as:

    \begin{equation}
        R(s) = \min \{r_{f}, r_{i} \times k^{s - 1}\} \label{eq:eq_resolution}
    \end{equation}

    To improve the use of available memory, the Adam optimizer~\cite{adam_optim}
    is used instead of the L-BFGS optimizer, which allows the processing of higher
    resolution images while still producing similar results.

    The algorithm's style perception can be improved to yield more
    expressive results by setting a different weight for each style layer.
    The layers used for stylization are the same as
    in Gatys' 2015 method: \emph{relu1\_1}, \emph{relu2\_1},
    \emph{relu3\_1}, \emph{relu4\_1} and \emph{relu5\_1}, and the weights assigned to
    respectively each layer is $256, 64, 16, 4$ and $1$, what is then
    normalized through the \emph{softmax} function.

    To approximate the effects of gradient normalization and produce
    better visual effects, a variation of the \emph{mean squared error} function
    is used to compute the content and style losses.
    Here, the traditional \emph{squared error} is divided by the sum of the
    absolute difference of the inputs, in a way that the gradient L1-norm of the resulting
    function will be $\approx 1$.
    For a given input $y$, a target $\hat y$ and an $\epsilon = 1 \mathrm{e}{-8}$
    value to avoid zero divisions, the \emph{normalized squared error} function is defined as:

    \begin{equation}
        NSE(y, \hat y) = \frac{
            \sum_{i} (y_{i} - \hat y_{i})^2
        }
        {
            \sum_{i} | y_{i} - \hat y_{i} | + \epsilon
        } \label{eq:eq_scaled_mse}
    \end{equation}

    The Gram Matrix for style representation was also changed.
    Now it is normalized by the number of filters contained
    in each style layer:

    \begin{equation}
        G^l_{ij} = \frac{\sum_{k} F^l_{ik} \times F^l_{jk}}{N_{l}} \label{eq:eq_gram_matrix_norm}
    \end{equation}

    Finally, following the paper \emph{Understanding Deep Image Representations by
    Inverting Them}~\cite{tv_loss}, spatial smoothness in the resulting image
    $\vec{x}$ can be encouraged by the \emph{L2 total variation} loss, defined as:

    \begin{equation}
        TV_{loss}(\vec{x}) = \frac{1}{N} \sum_{i,j}^{N} \Bigl( (\vec{x}_{i, j + 1} - \vec{x}_{ij})^2 + (\vec{x}_{i + 1, j} - \vec{x}_{ij})^2 \Bigr) \label{eq:eq_tv_loss}
    \end{equation}

    So, it is summed with the \emph{content} and \emph{style} losses with
    a weight control value $\gamma$, defining the final \emph{total loss} as:

    \begin{equation}
        \mathcal{L}_{total}(\vec{c}, \vec{s}, \vec{x}) = \alpha \mathcal{L}_{content}(\vec{c}, \vec{x}) + \beta \mathcal{L}_{style}(\vec{s}, \vec{x}) + \gamma TV_{loss}(\vec{x}) \label{eq:eq_total_crowson}
    \end{equation}

\begin{figure}[htp]
        \centering
        \subfloat[Gatys et al.]{
            \includegraphics[width=2.3in]{assets/gatys_samples/golden_gate_painting_gatys_2016}
        }
        \hspace{.1em}%
        \subfloat[K. Crowson]{
            \includegraphics[width=2.3in]{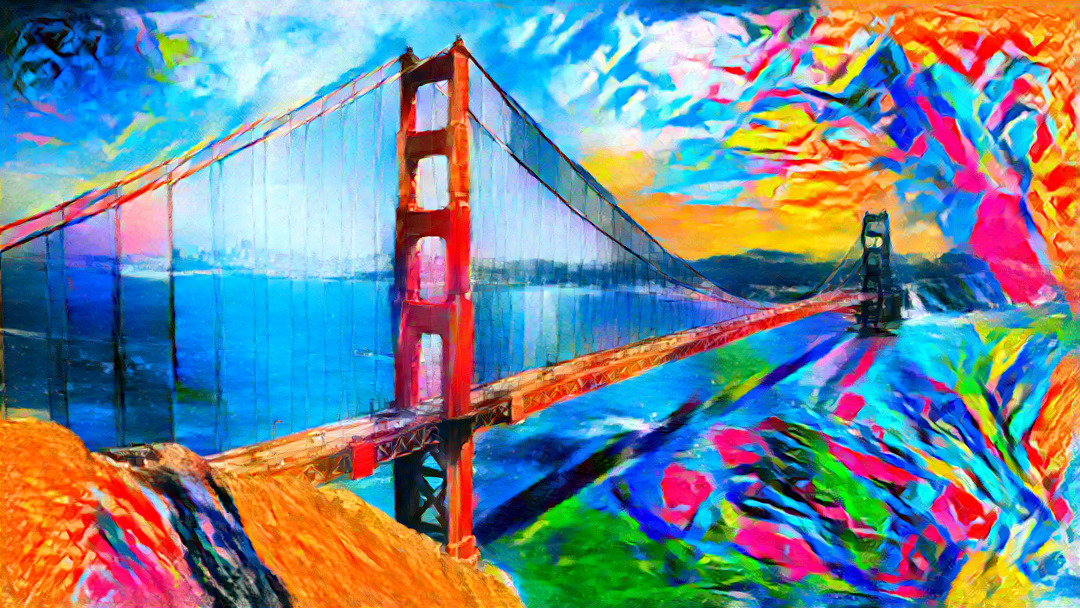}
        }
    
        \caption{
            Comparison between the Gatys' 2016 style transfer method
            before (a) and after (b) Crowson's changes. After these improvements,
            the image content becomes well-preserved while the stylized regions are
            more expressive.
        }
        \label{fig:crowson_and_gatys}
    \end{figure}

    With all these changes and a few others that can be found in Crowson's repository,
    the results of the style transfer process are even improved, resulting in
    smoother images, with well-preserved content and more expressive strokes
    (Figure~\ref{fig:crowson_and_gatys}(b)).

    \subsection{Limitations} \label{subsec:limitations}

\begin{figure}[htp]
            \centering
            \subfloat[Content Image]{
                \includegraphics[height=1.3in]{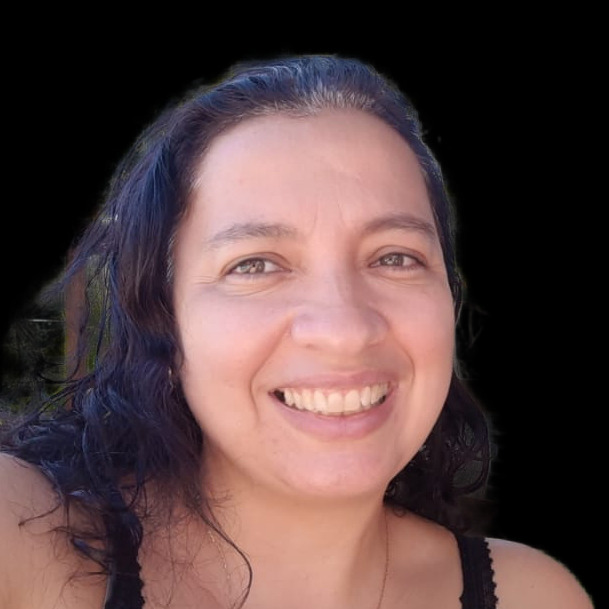}
            }
            \hspace{.1em}%
            \subfloat[Style Image]{
                \includegraphics[height=1.3in]{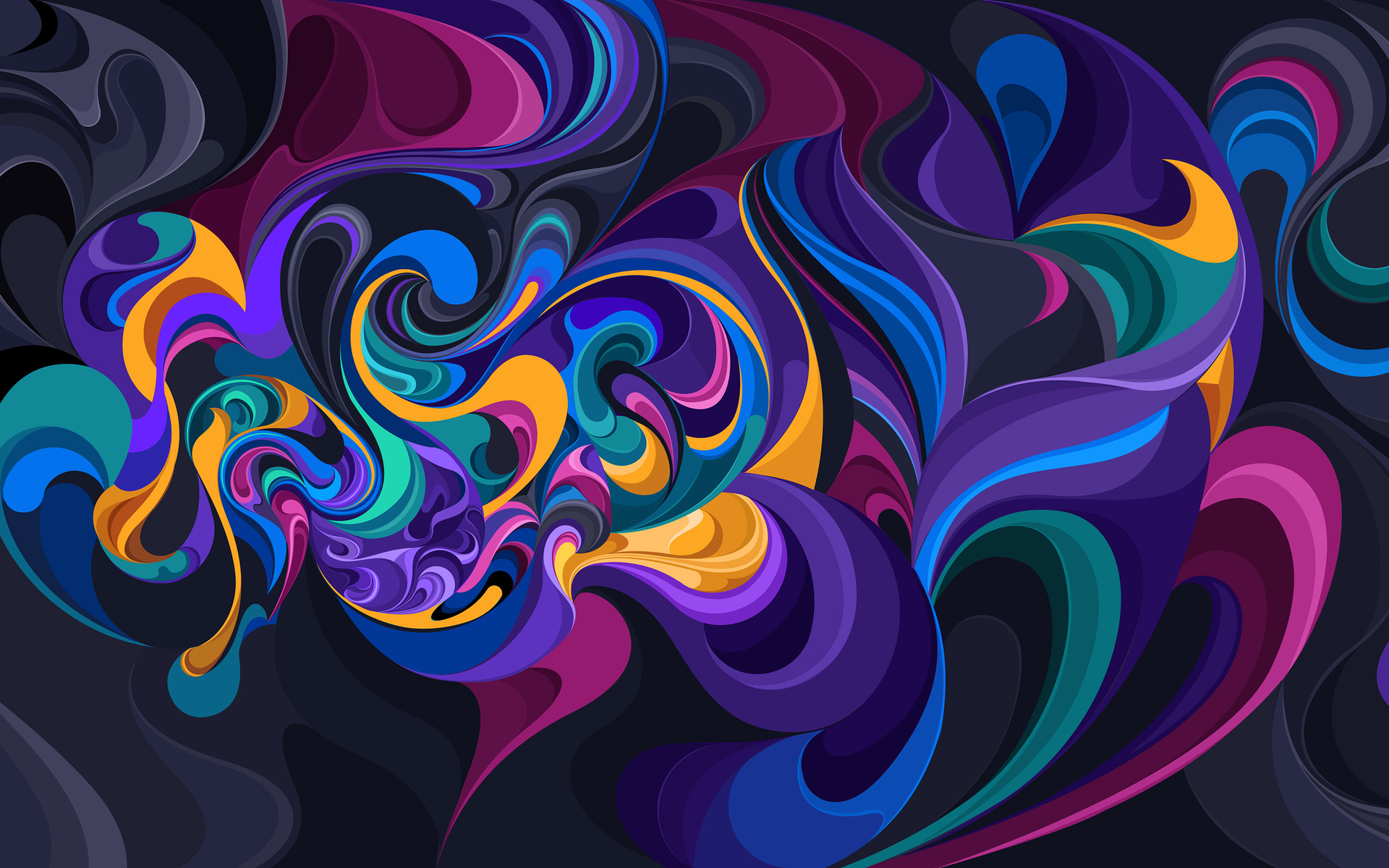}
            }
            \hspace{.1em}%
            \subfloat[K. Crowson]{
                \includegraphics[height=1.3in]{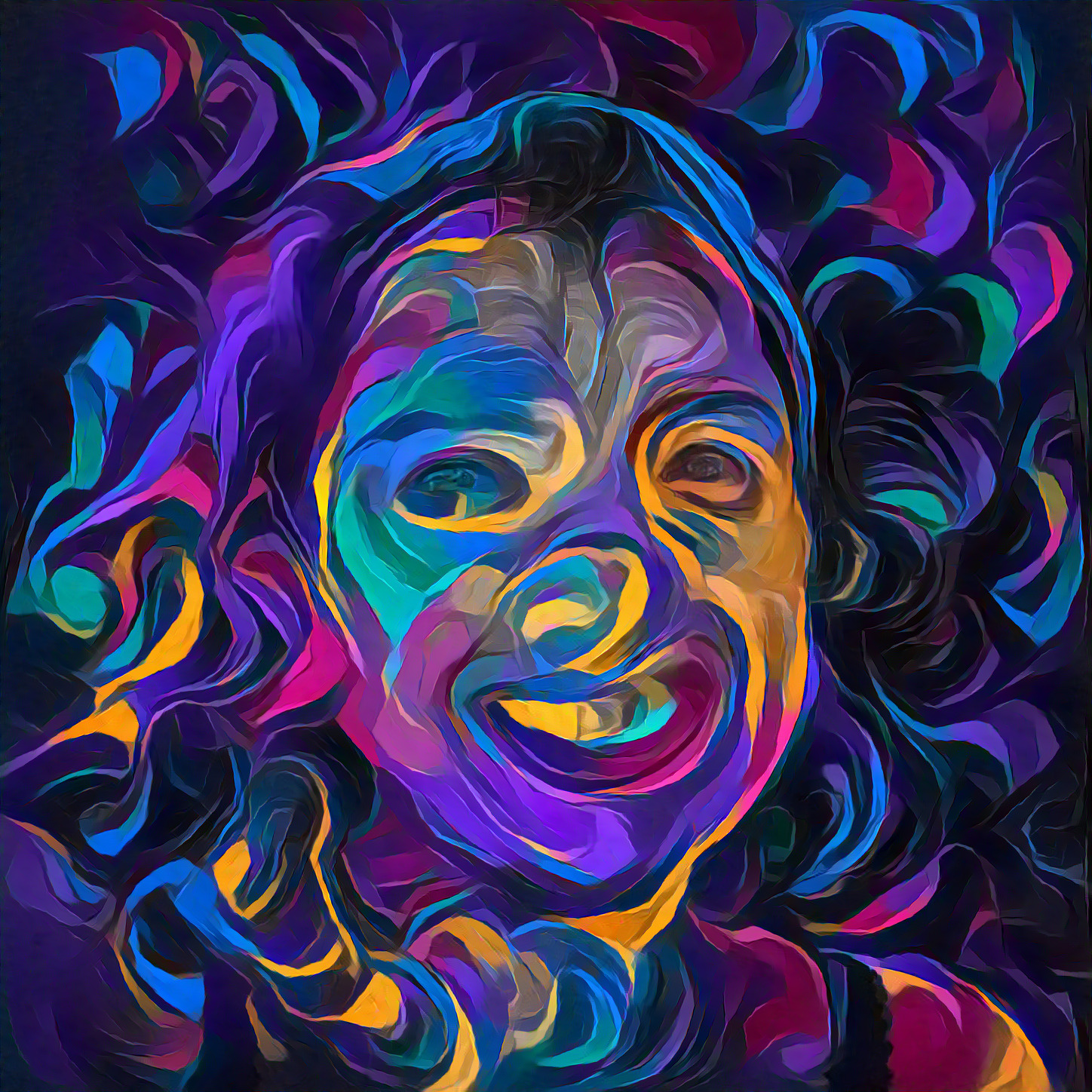}
            }
        
            \caption{
                A failure case of the Crowson's method, where the facial features of the
                \emph{content image} are not propagated to the final \emph{result image},
                making the face contained in the result unrecognizable.
            }
            \label{fig:crowson_fail_case}
        \end{figure}

        Despite those improvements, the method still produces facial deformations
        when applied to content images with human faces
        (Figure~\ref{fig:crowson_fail_case}(c)), as the content layer \emph{relu4\_2}
        can't output meaningful representations of human facial features.
    
        These distortions in the generated results (Figure~\ref{fig:crowson_fail_case}(c))
        occurs because the VGG-Network layers used for content extraction (\emph{relu4\_2}
        in the Gatys' and Crowson's methods)
        can't output meaningful activations about human faces when exposed to it,
        as this network was trained for the general-purpose image classification task
        and not for domain-specific tasks like face recognition.

\section{Proposed Method} \label{sec:method}

    This paper proposes a new method, that will be called as
    \emph{Portrait Stylization}, to solve the face distortion problem in Crowson's algorithm.
    It solves the problem by adding to the \emph{total loss} function
    a new domain-specific content loss, called here as \emph{FaceID loss}, that
    uses the responses from the convolutional filters of a pre-trained face
    recognition algorithm, like the state-of-the-art FaceNet~\cite{facenet}, to
    compute the difference between the facial features of the \emph{content image}
    $\vec{c}$ and the \emph{result image} $\vec{x}$.

    These responses are extracted from an \emph{Inception-Resnet-V1} FaceNet model,
    pretrained on the \emph{VGGFace2}~\cite{vggface2} dataset, and will be called
    here as \emph{facial features}.
    The layers selected for extracting these \emph{facial features} are:
    \emph{conv\_1a}, \emph{conv\_2a}, \emph{maxpool\_3a}, \emph{conv\_4a}
    and \emph{conv\_4b}, with the same weight assigned to each one.
    It was empirically selected, following the idea that higher layers
    in a feed-forward convolutional neural network architecture are better in
    extracting general-features.

    The \emph{FaceID loss} function can be defined as the weighted sum
    of the \emph{normalized squared error} (Eq.~\ref{eq:eq_scaled_mse})
    between the \emph{content image facial features} $C^l$ and the
    \emph{result image facial features} $X^l$, being $W_{l}^{f}$ the
    weight for a given FaceNet layer $l$:

    \begin{equation}
        \mathcal{L}_{facial}(\vec{c}, \vec{x}) = \delta \times \sum_{l=0}^{L} W_{l}^{f} \times NSE(C^l, X^l)  \label{eq:eq_faceid}
    \end{equation}

    Where $\delta$ is the \emph{face weight} value, that gives the user control
    of how similar the faces contained in the \emph{result image} will be
    in comparison to the faces in the \emph{content image}.

\begin{figure}[htp]
        \centering
        \subfloat[K. Crowson]{
            \includegraphics[width=1.2in]{assets/grid/1/crowson}
        }
        \hspace{.1em}%
        \subfloat[P.S. Method]{
            \includegraphics[width=1.2in]{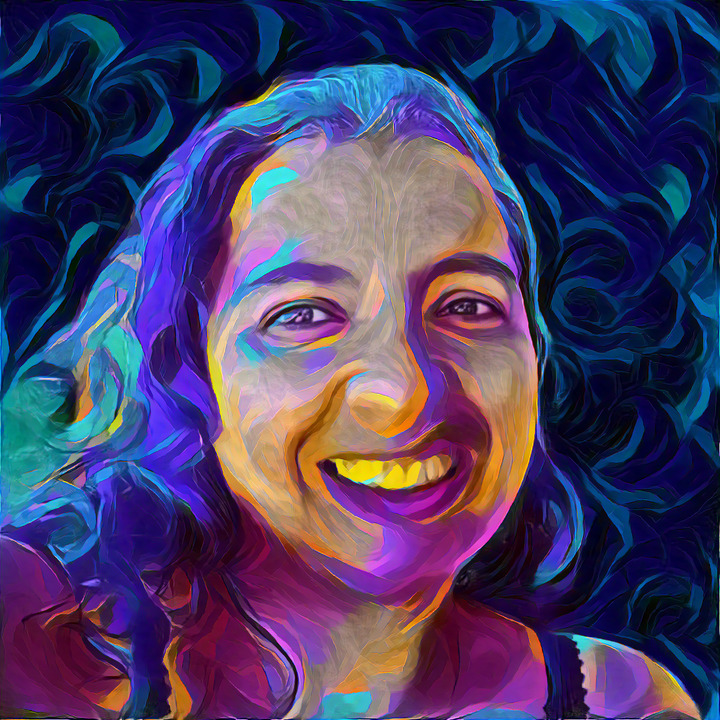}
        }
    
        \caption{
            Comparison between the Crowson's style transfer method
            before (a) and after (b) the addition of the \emph{FaceID Loss}.
        }
        \label{fig:face_loss_example}
    \end{figure}

    This auxiliary criterion helps the algorithm retain the facial
    features of the human faces in the \emph{content image} after the stylizing process,
    which avoids drastic facial deformations while still producing expressive stylization
    results (Figure~\ref{fig:face_loss_example}(b)).

\begin{figure}[htp]
        \centering
        \subfloat[$\delta = 0.05$]{
            \includegraphics[width=1.in]{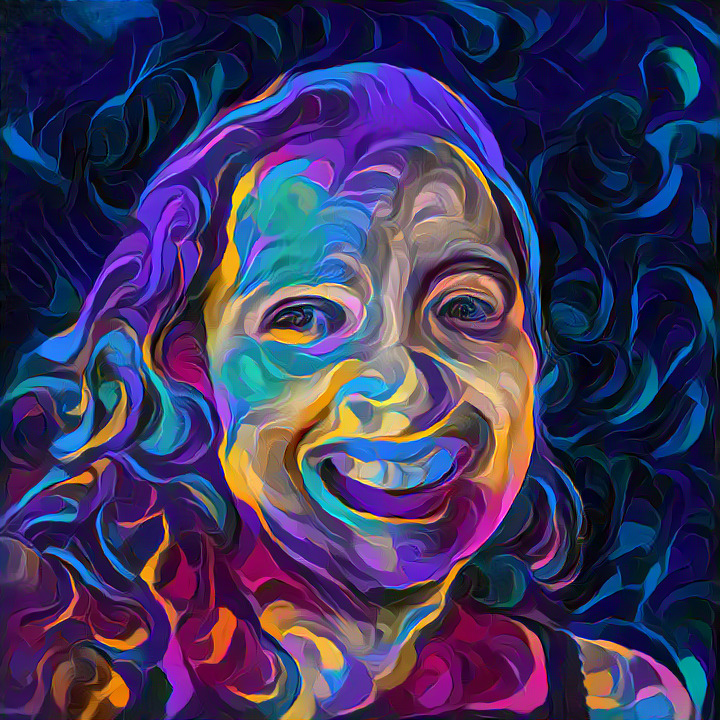}
        }
        \hspace{.1em}%
        \subfloat[$\delta = 0.15$]{
            \includegraphics[width=1.in]{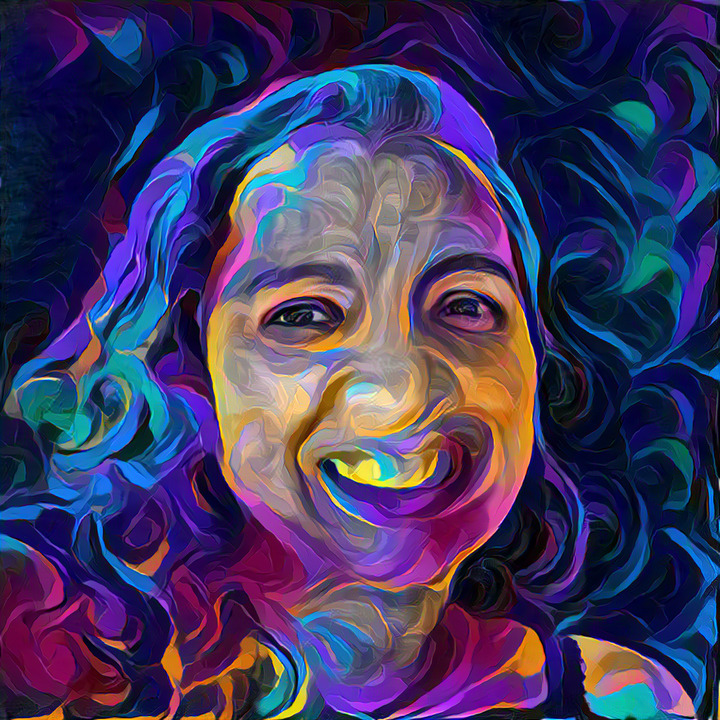}
        }
        \hspace{.1em}%
        \subfloat[$\delta = 0.25$]{
            \includegraphics[width=1.in]{assets/mine_samples/faceloss_spectrum/fw_25}
        }
        \hspace{.1em}%
        \subfloat[$\delta = 0.35$]{
            \includegraphics[width=1.in]{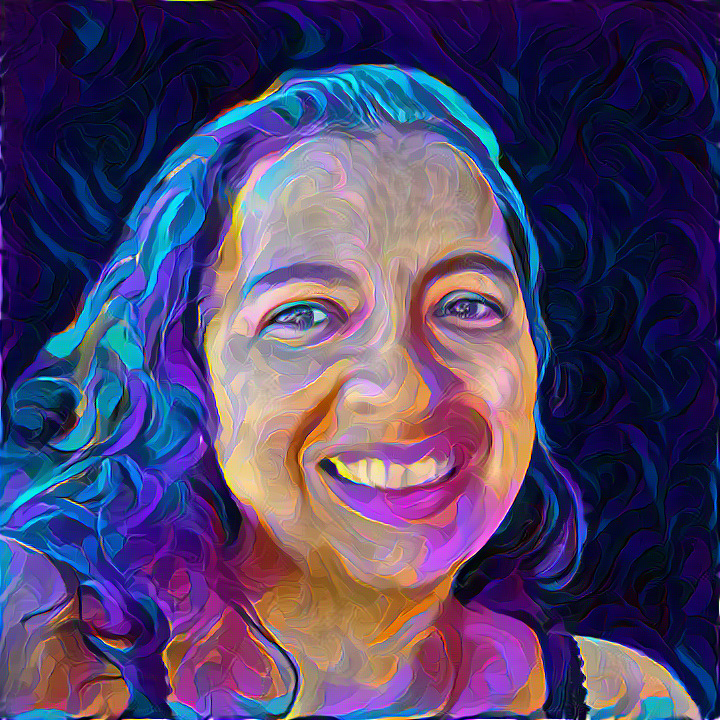}
        }
        \hspace{.1em}%
        \subfloat[$\delta = 0.45$]{
            \includegraphics[width=1.in]{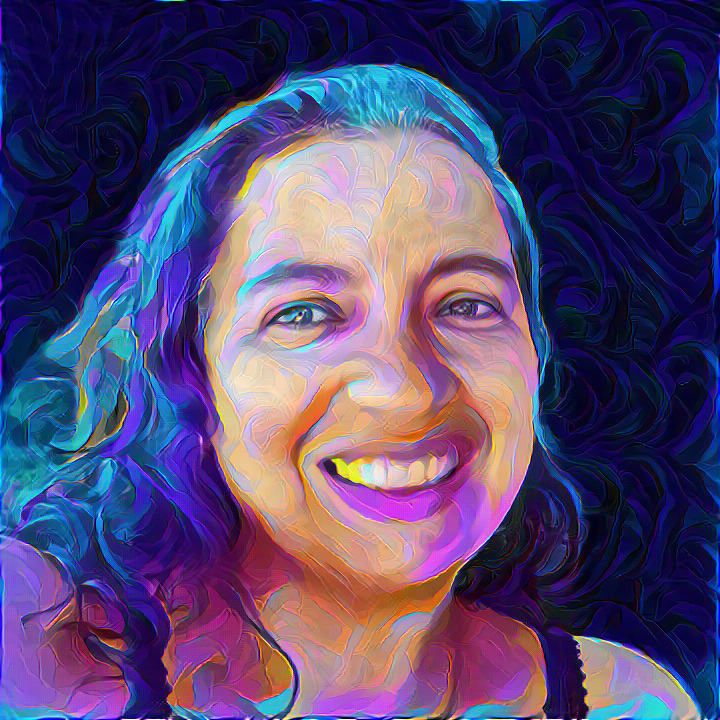}
        }
    
        \caption{
            Changes caused by the \emph{FaceID loss}, from $\delta = 0.05$
            until $\delta = 0.45$, with a step size of $\Delta = 0.1$.
            All these experiments use $\alpha = 0.05$ and $\beta = 1.0$,
            with the same \emph{content} and \emph{style} images from
            Figure~\ref{fig:crowson_fail_case}.
        }
        \label{fig:face_loss_spectrum}
    \end{figure}

    Nevertheless, even with those improvements, using only the \emph{facial features}
    as a criterion for the face reconstruction don't give the user much control about
    the style of \emph{result image}, as the style strokes becomes lower expressive
    as the $\delta$ value increases (Figure~\ref{fig:face_loss_spectrum}).

    \Needspace{10\baselineskip}

    In that way, to improve the control of the face geometry in the
    \emph{result image} while maintaining the expressive stylization
    results, the differentiable output from the FaceMesh~\cite{facemesh}
    algorithm can be used to compute the difference between the surface
    geometry of the faces contained in the \emph{content} and \emph{result}
    images.

\begin{figure}[htp]
        \centering
        \subfloat[]{
            \includegraphics[width=.9in]{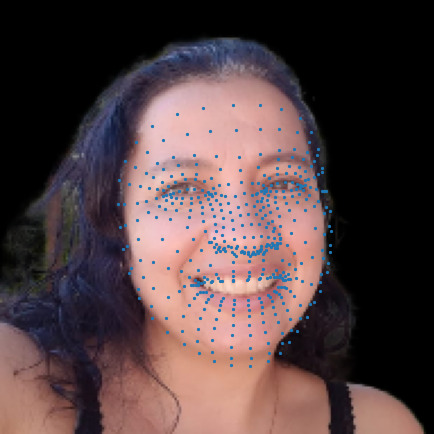}
        }
        \hspace{.1em}%
        \subfloat[]{
            \includegraphics[width=.9in]{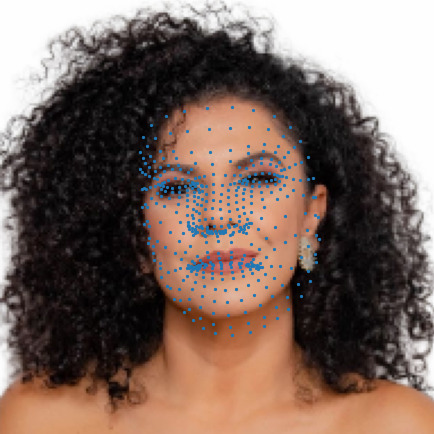}
        }
        \hspace{.1em}%
        \subfloat[]{
            \includegraphics[width=.9in]{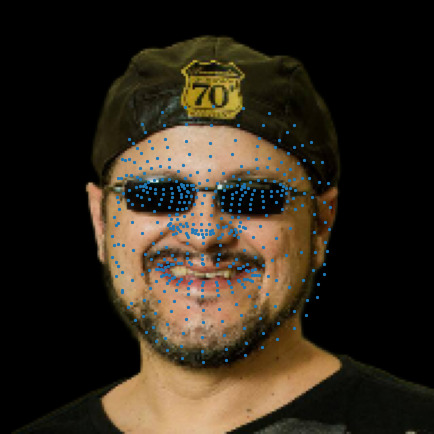}
        }
    
        \caption{
            Examples of \emph{mesh models} generated by the FaceMesh algorithm
            applied to \emph{content images} with real human faces. To produce
            high-precision meshes, the faces on the input image need to be cropped
            and resized to 192x192 resolution.
        }
        \label{fig:face_mesh_example}
    \end{figure}

    The FaceMesh algorithm is a feed-forward convolutional neural network
    trained to approximate a 3D mesh representation of human faces from
    only RGB image inputs (Figure~\ref{fig:face_mesh_example}).
    It outputs a relatively dense \emph{mesh model} of 468 vertices
    that was designed for face-based \emph{augmented reality} effects.

    So, the new \emph{FaceID loss} can be defined by adding to the old one
    the \emph{normalized squared error} (Eq.~\ref{eq:eq_scaled_mse}) between
    the \emph{content mesh model} $C_{M}$ and the \emph{result mesh model} $X_{M}$:

    \begin{equation}
        \mathcal{L}_{facial}(\vec{c}, \vec{x}) =
            \eta NSE(C_{M}, X_{M}) +
            \delta \sum_{l=0}^{L} W_{l}^{f} \times NSE(C^l, X^l)
        \label{eq:eq_facemesh}
    \end{equation}

    Where $\eta$ is the \emph{meshes weight} value, that gives the user control
    of how much similar the surface geometry of faces contained in the
    \emph{result image} will be in relation to the ones in the \emph{content image}.

    This encourages the algorithm to reproduce the face geometries contained
    on the \emph{content image}, on the \emph{result image}, while still allows it
    to produce expressive strokes, as the other facial features like
    the skin texture or micro expressions is not represented by the \emph{mesh model}.

    \begin{figure}[htp]
        \centering
        \subfloat[$\delta = 0.00, \eta = 0.00$]{
            \includegraphics[width=1.3in]{assets/grid/1/crowson}
        }
        \hspace{.1em}%
        \subfloat[$\delta = 0.00, \eta = 0.015$]{
            \includegraphics[width=1.3in]{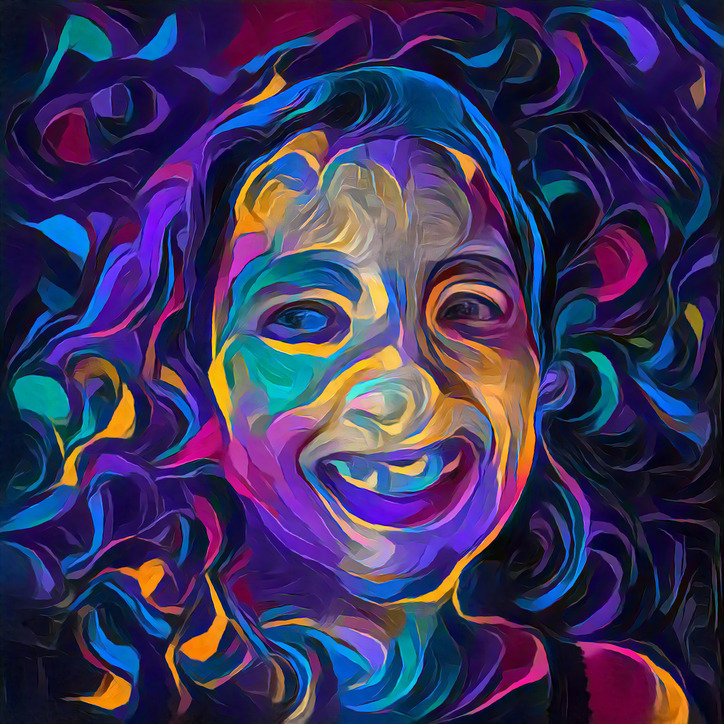}
        }
        \hspace{.1em}%
        \subfloat[$\delta = 0.25, \eta = 0.00$]{
            \includegraphics[width=1.3in]{assets/mine_samples/faceloss_spectrum/fw_25}
        }
        \hspace{.1em}%
        \subfloat[$\delta = 0.25, \eta = 0.015$]{
            \includegraphics[width=1.3in]{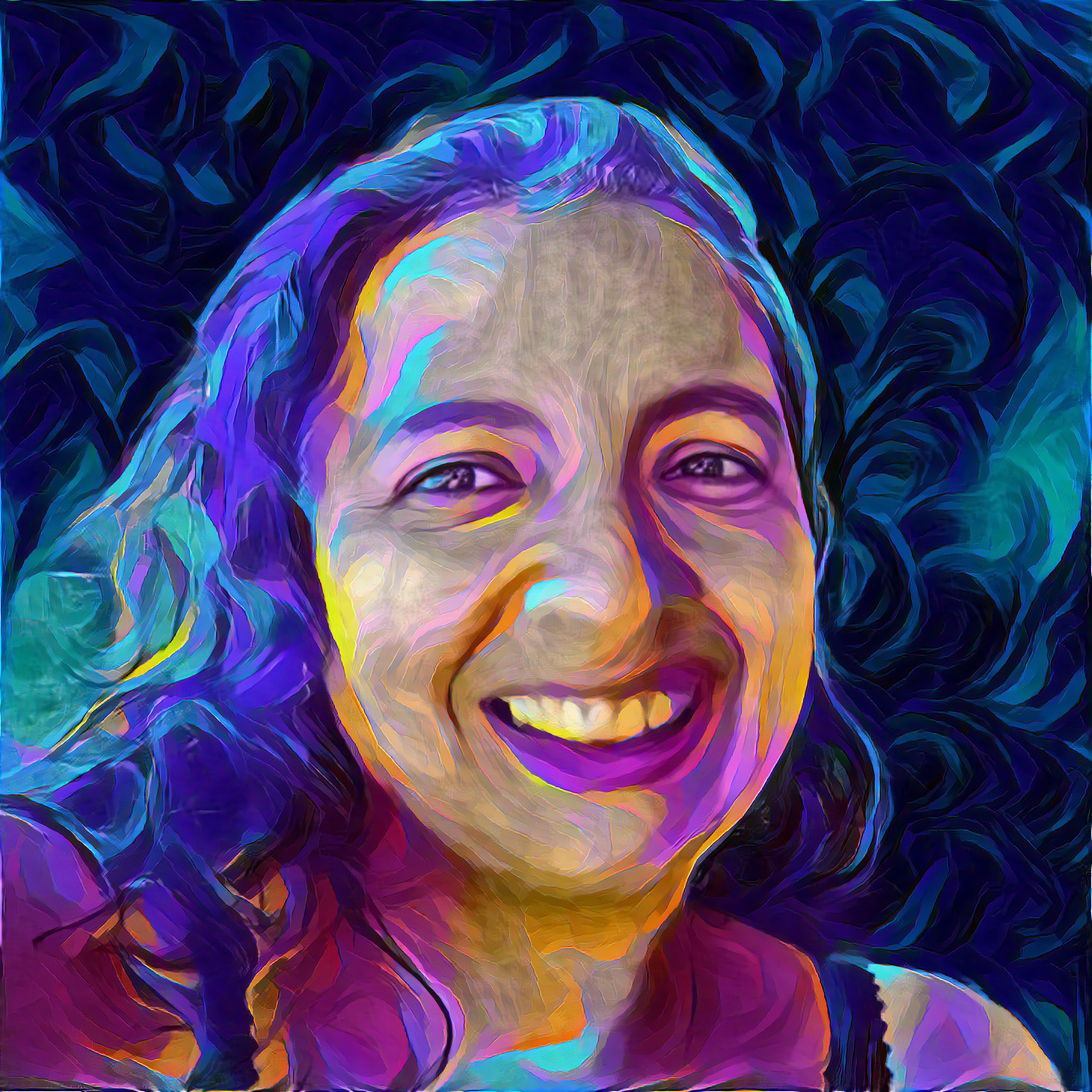}
        }
    
        \caption{
            Changes caused by the new \emph{FaceMesh term} in different
            algorithm settings. All these experiments use
            $\alpha = 0.05$ and $\beta = 1.0$, with the same \emph{content}
            and \emph{style} images from Figure~\ref{fig:crowson_fail_case}.
        }
        \label{fig:face_mesh_results}
    \end{figure}
    
    This new term improves the quality of face reconstructions in
    the \emph{result image} even when it is used alone.
    When used jointly with the \emph{facial features}
    (with $\delta > 0.00$), makes the algorithm produce even
    better results, helping the user to adjust fine facial geometry
    details like the nose or mouth formats, producing facial
    expressions most similar to the ones in the \emph{content image}
    (Figure~\ref{fig:face_mesh_results}(d)).

\section{Image Preprocessing} \label{sec:image-preprocessing}

\begin{figure}[htp]
        \centering
        \subfloat[Original]{
            \includegraphics[width=1.2in]{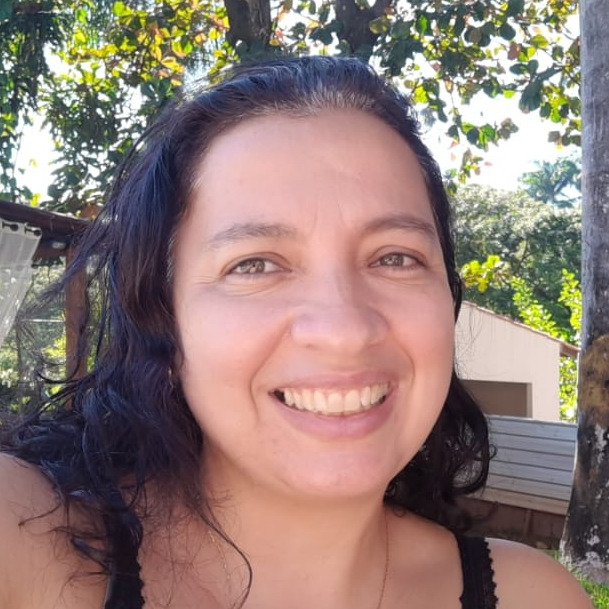}
        }
        \hspace{.1em}%
        \subfloat[Mask]{
            \includegraphics[width=1.2in]{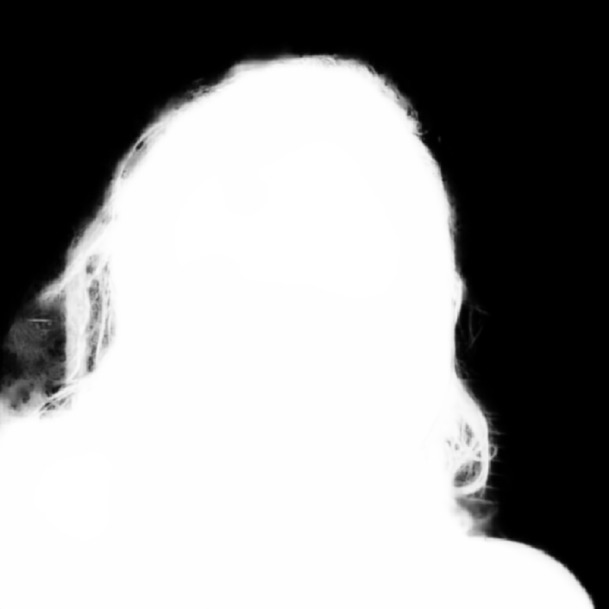}
        }
        \hspace{.1em}%
        \subfloat[Result]{
            \includegraphics[width=1.2in]{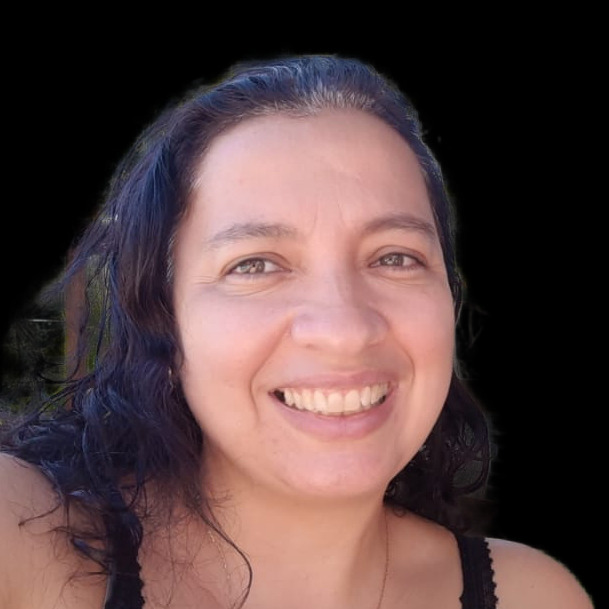}
        }
    
        \caption{
            An example of a \emph{content image} before (a) and after (c)
            the background removal process. All the black pixels in the mask (b)
            are replaced by a black color in the resulting image, and the white pixels
            are maintained with the same color values.
        }
        \label{fig:modnet_example}
    \end{figure}

    To improve the stylization results of the algorithms used here,
    making it focus only on the faces contained in the \emph{content image}
    and avoiding possible confusion caused by the background texture,
    the background was removed and replaced by a flat color through
    the MODNet Trimap-Free Portrait Matting~\cite{modnet} algorithm
    (Figure~\ref{fig:modnet_example}).

\begin{figure}[htp]
        \centering
        \subfloat[Gatys et al.]{
            \includegraphics[width=1.2in]{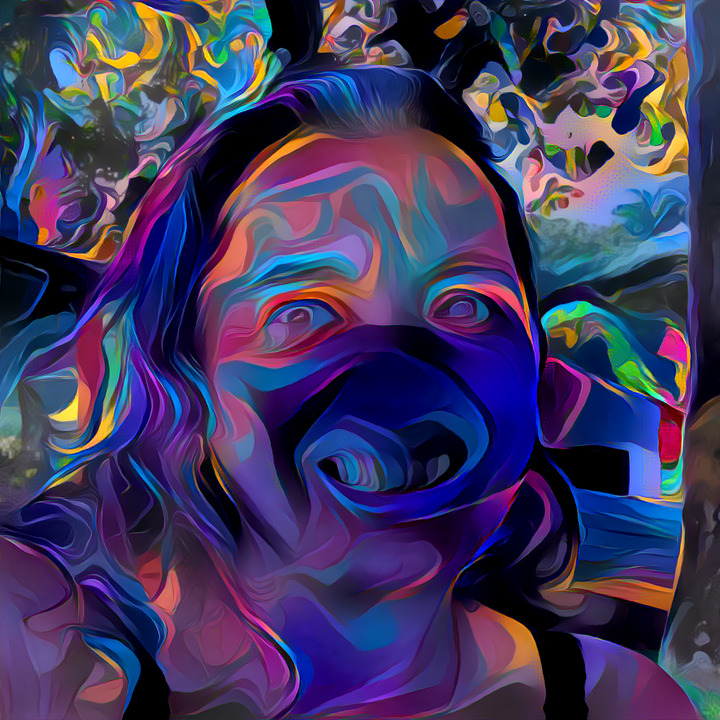}
        }
        \hspace{.1em}%
        \subfloat[K. Crowson]{
            \includegraphics[width=1.2in]{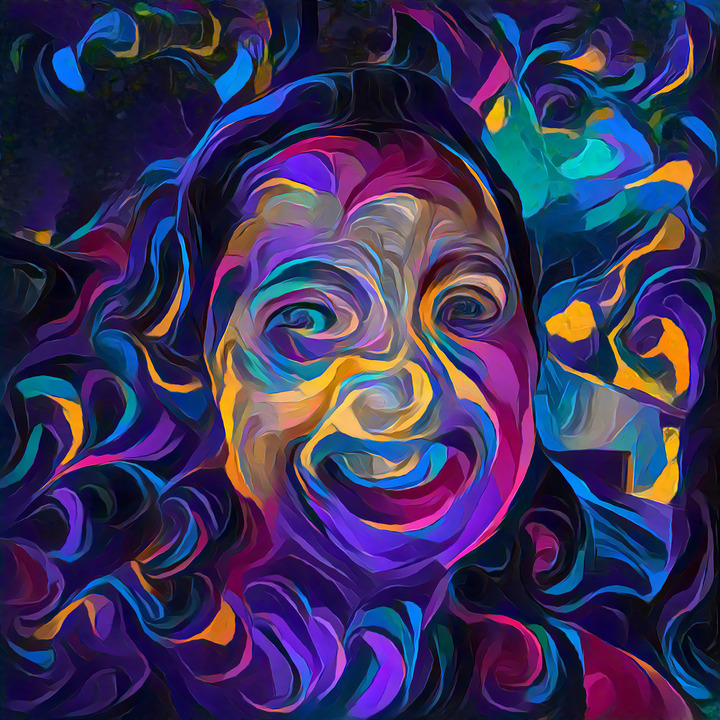}
        }
        \hspace{.1em}%
        \subfloat[P.S. Method]{
            \includegraphics[width=1.2in]{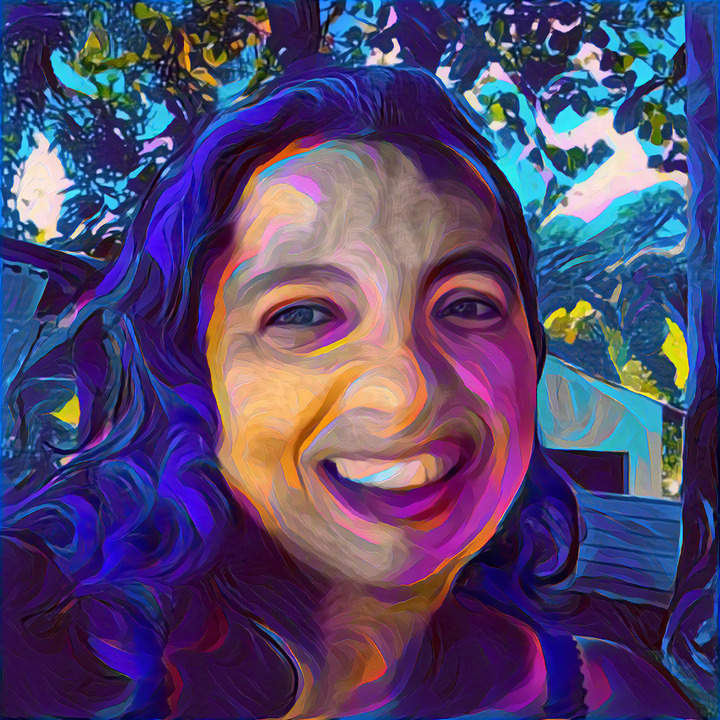}
        }
    
        \caption{
            The results of stylizing a portrait image without
            removing its background. To generate these samples,
            the same \emph{style image} and hyperparameters from
            Figure~\ref{fig:face_mesh_results}(d) were used.
        }
        \label{fig:background_stylization_example}
    \end{figure}

    This step is not needed by the algorithms, but it improves
    their overall performance in portrait images.
    The \emph{Portrait Stylization} method can generate an image
    with only a few facial distortions even with its background,
    but the other algorithms seem to yields bigger distortions
    when applied to this same case
    (Figure~\ref{fig:background_stylization_example}).

\Needspace{10\baselineskip}

\section{Style Transfer with Multiple Face Portraits} \label{sec:style-transfer-on-multiple-faces}

\begin{figure}[htp]
        \centering
        \subfloat[Content Image]{
            \includegraphics[width=2.in]{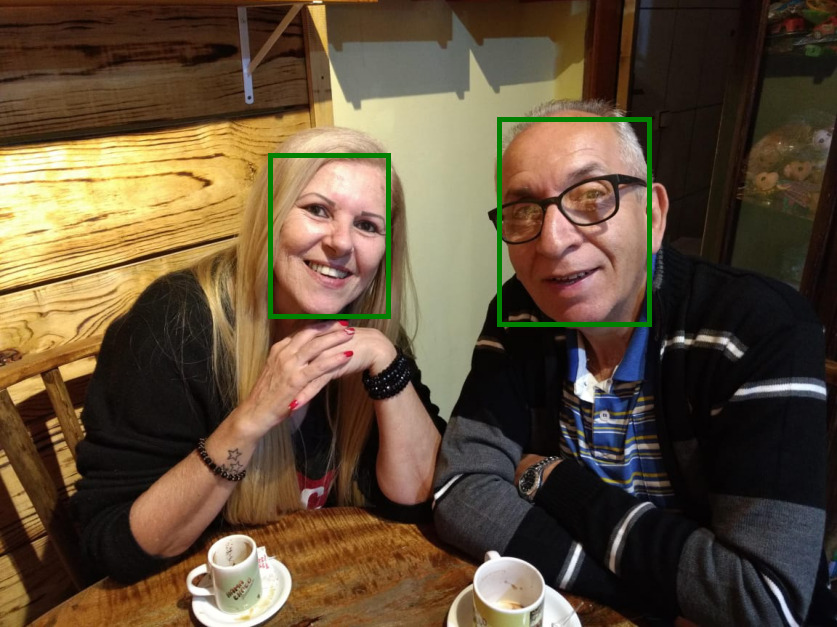}
        }
        \hspace{.1em}%
        \subfloat[Result Image]{
            \includegraphics[width=2.in]{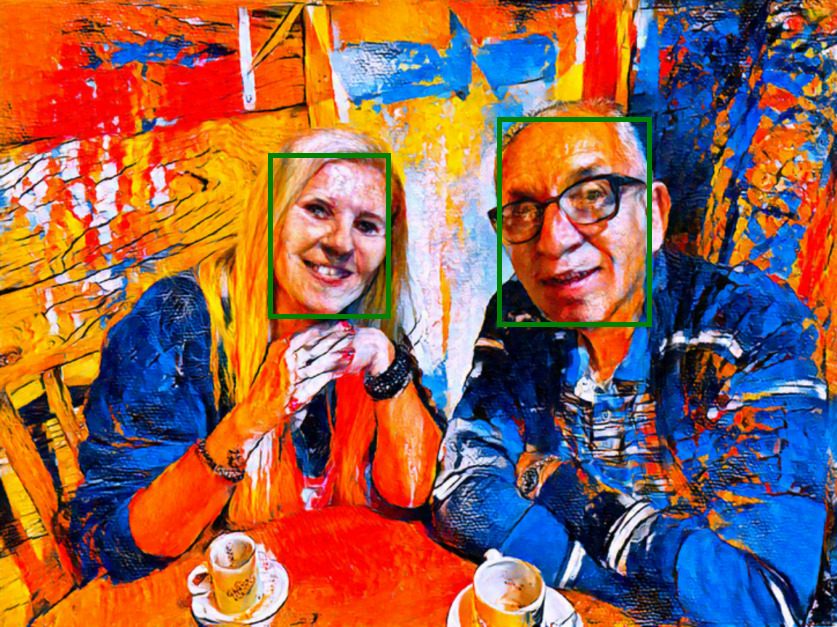}
        }
    
        \caption{
            An example of how the MTCNN algorithm extracts the face coordinates
            from the \emph{content image} and uses them to crop the faces in the
            \emph{result image}. This process ensures that the faces in the
            generated image will always be detected, no matter what changes
            the styling process does.
        }
        \label{fig:mtcnn_example}
    \end{figure}

    To enable the support of stylizing images that contain more than
    one human face, a few changes need to be made to the algorithm.
    A face detection algorithm like the state-of-the-art MTCNN~\cite{mtcnn}
    is used to extract the coordinates of the faces contained in the \emph{content image}.
    With this coordinates, the facial regions in the \emph{content} and \emph{result}
    images are cropped, generating sub-images from the ground truth and generated faces
    (Figure~\ref{fig:mtcnn_example}).
    
    The \emph{facial features} of all these sub-images are extracted through the
    FaceNet and FaceMesh models.
    The features from the \emph{content image} faces are concatenated layer
    by layer, generating a tuple of vectors $G$ that represents the ground truth
    facial geometries of all faces.
    At the same time, the extracted features from the \emph{result image} faces are too
    concatenated layer by layer, generating a tuple of vectors $H$ that represents the
    actual geometries of the faces in the generated image.

    After these processes, the \emph{FaceID Loss} are calculated using these concatenated vectors
    as facial representations rather than using the \emph{facial features} extracted from the
    entire image pixels.

    \begin{equation}
        \mathcal{L}_{facial}(\vec{c}, \vec{x}) =
            \eta NSE(G_{M}, H_{M}) +
            \delta \sum_{l=0}^{L} W_{l}^{f} \times NSE(G_{f}^l, H_{f}^l)
        \label{eq:eq_multi_face}
    \end{equation}

    Where $G_{M}$ and $G_{f}^l$ are the concatenated \emph{mesh models}
    and \emph{facial features} from the faces in the \emph{content image},
    and $H_{M}$ and $H_{f}^l$ are the concatenated \emph{mesh models}
    and \emph{facial features} from the faces in the \emph{result image}.

\Needspace{10\baselineskip}

\section{Performance Comparisons} \label{sec:performance-comparisons}

    The \emph{Portrait Stylization} method produces significant visual
    improvements, compared to the current state-of-the-art methods, when
    applied to images that contain human faces (Table~\ref{tab:table_comparisons}).
    This still can be applied to any other image types as the method becomes
    exactly the same as Crowson's method when $\delta = 0.00$ and $\eta = 0.00$.

    \newcommand{\putimage}[1]{\includegraphics[width=1.in]{assets/grid/#1}}
    \newcolumntype{C}{>{\centering\arraybackslash}m{6em}}

    \begin{table}[htb]
        \centering
        \begin{tabular}{l*4{C}@{}}
            \toprule
            \# & Input & Gatys et al.\ & K. Crowson & P.S.\ Method \\
            \midrule
            1 & \putimage{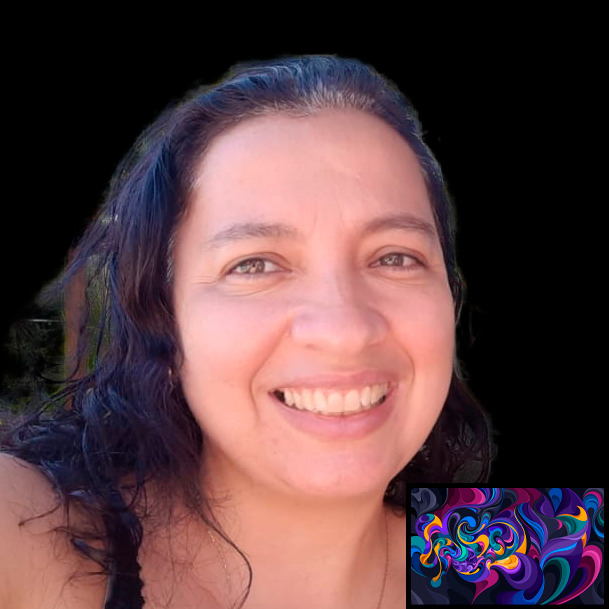} & \putimage{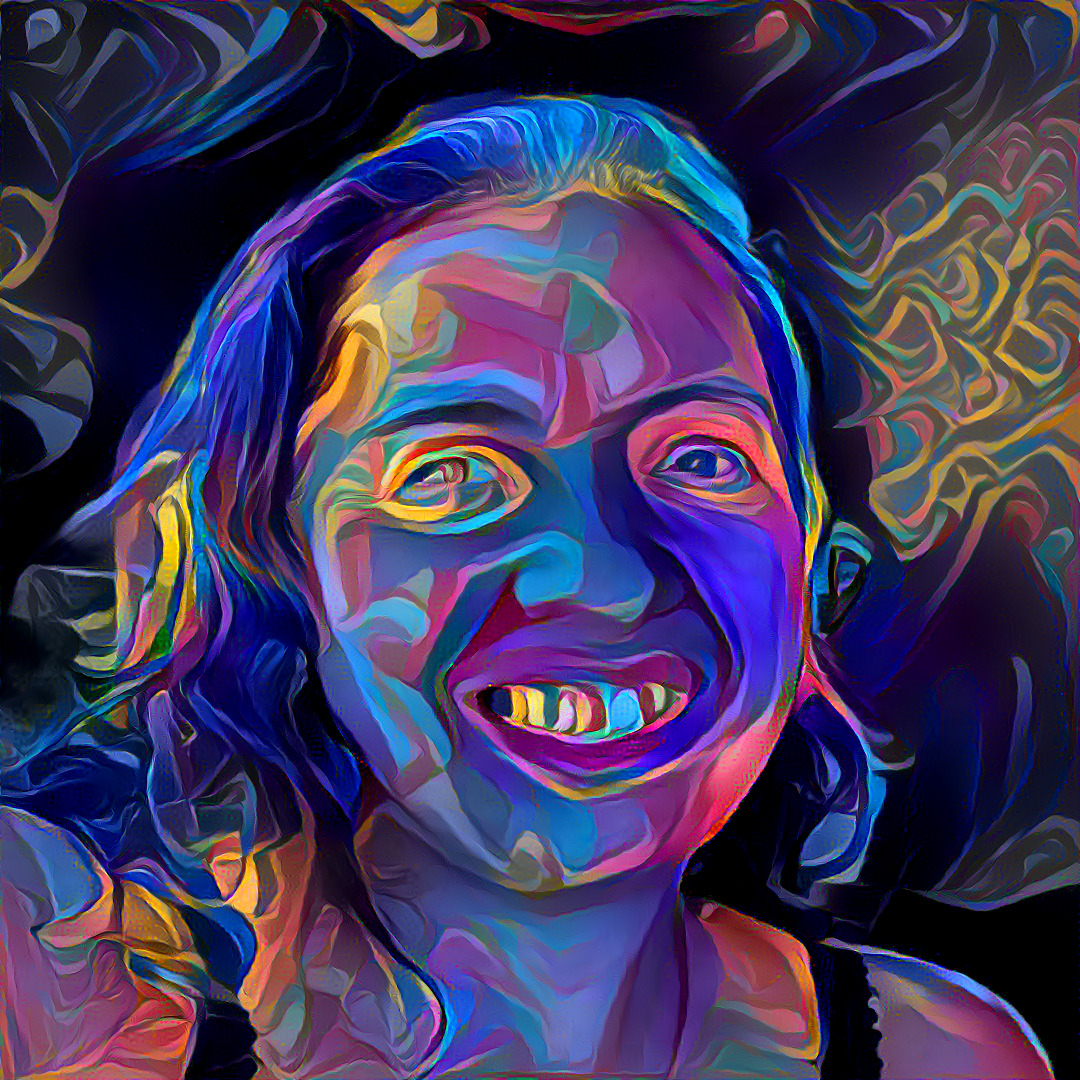} & \putimage{1/crowson} & \putimage{1/this} \\
            2 & \putimage{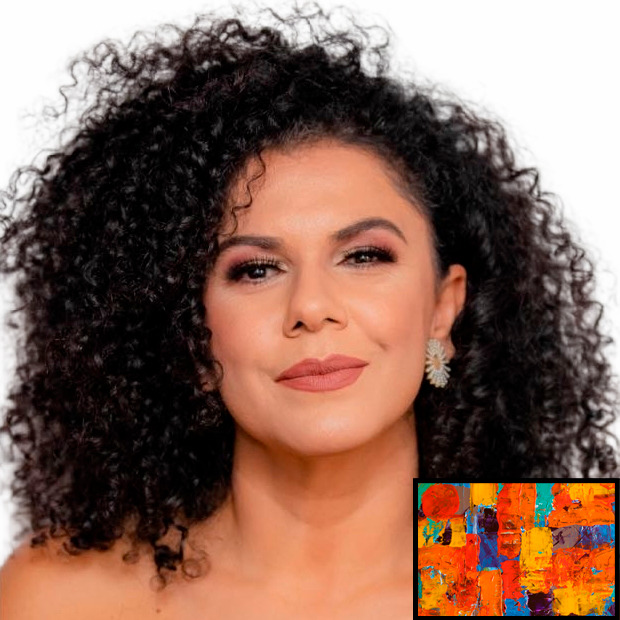} & \putimage{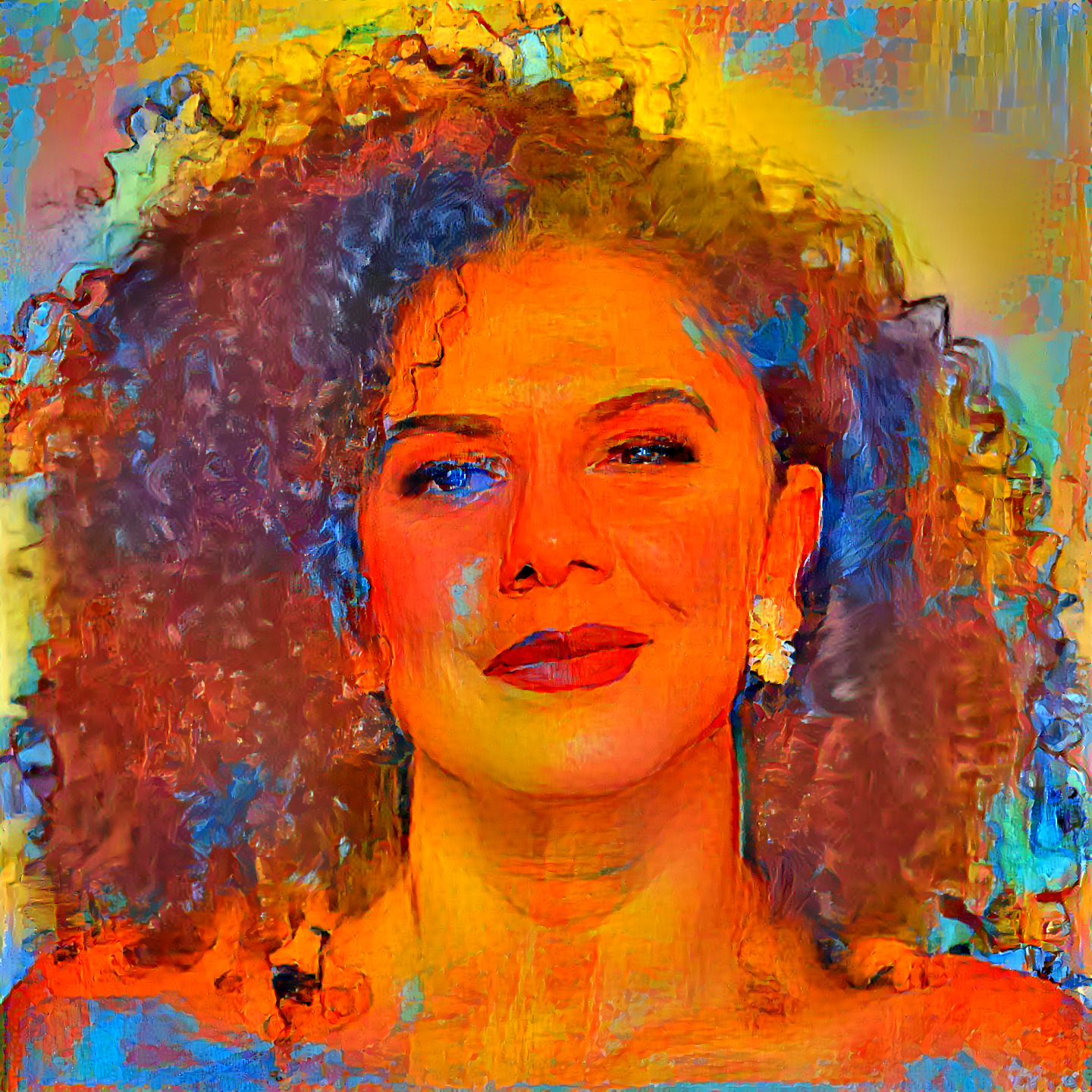} & \putimage{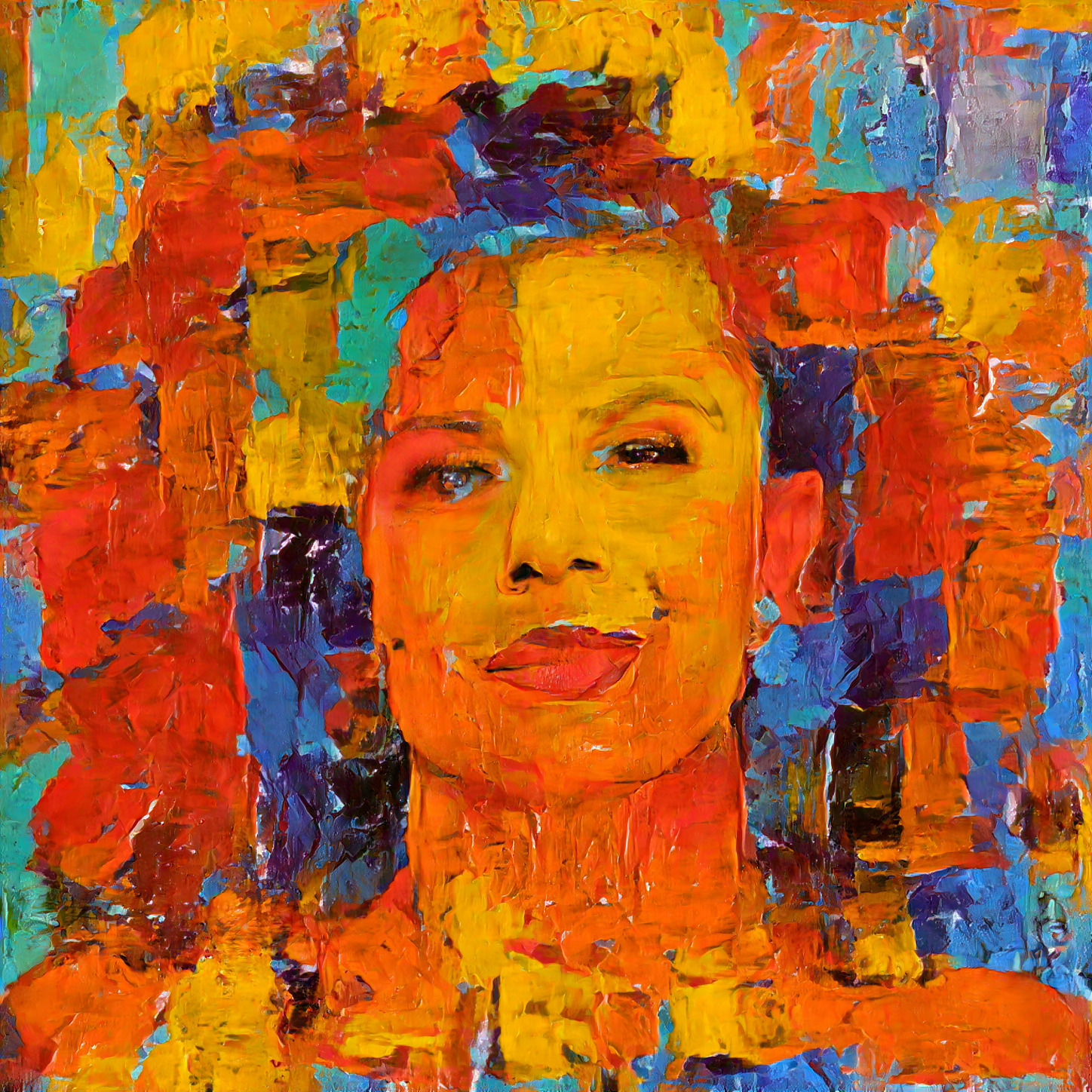} & \putimage{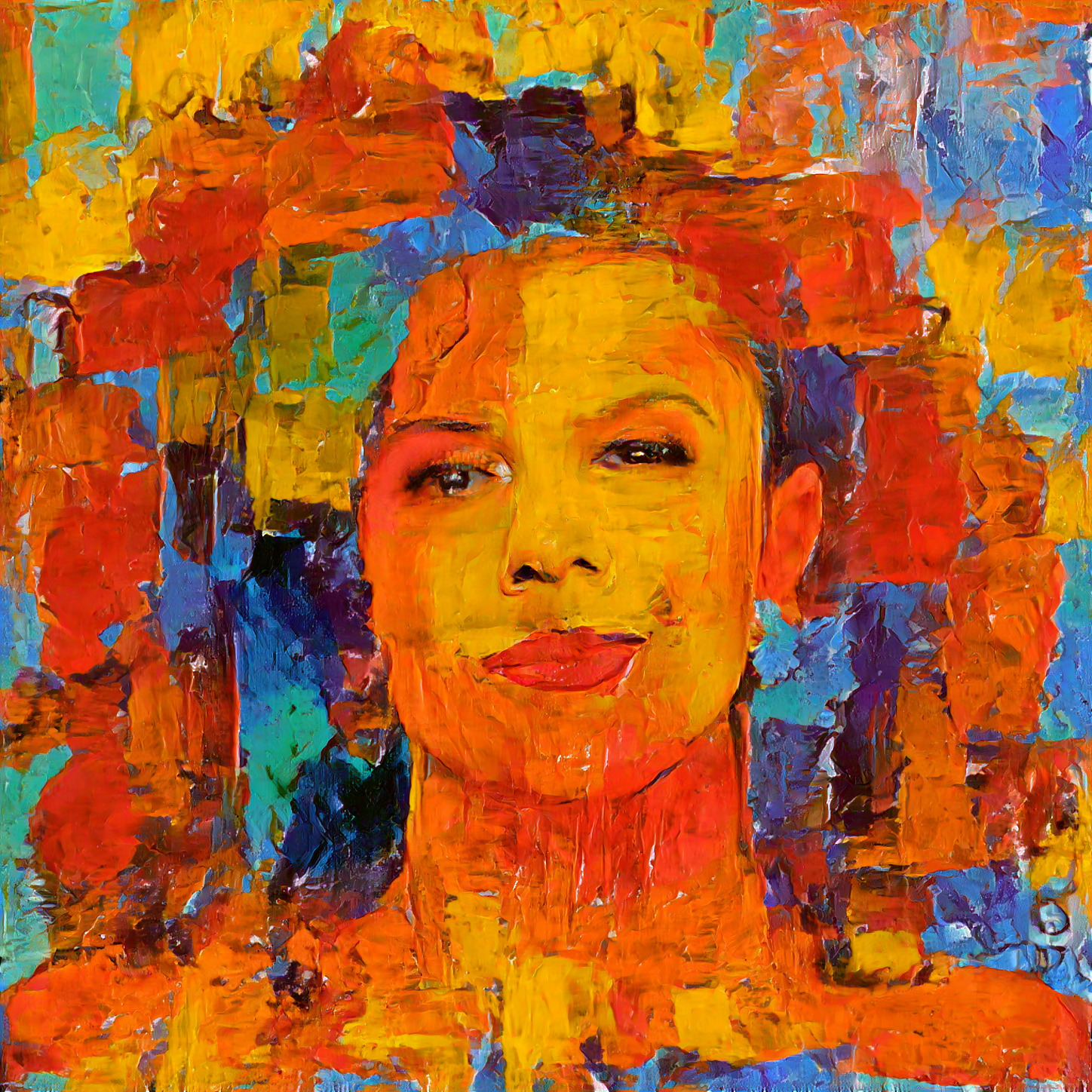} \\
            3 & \putimage{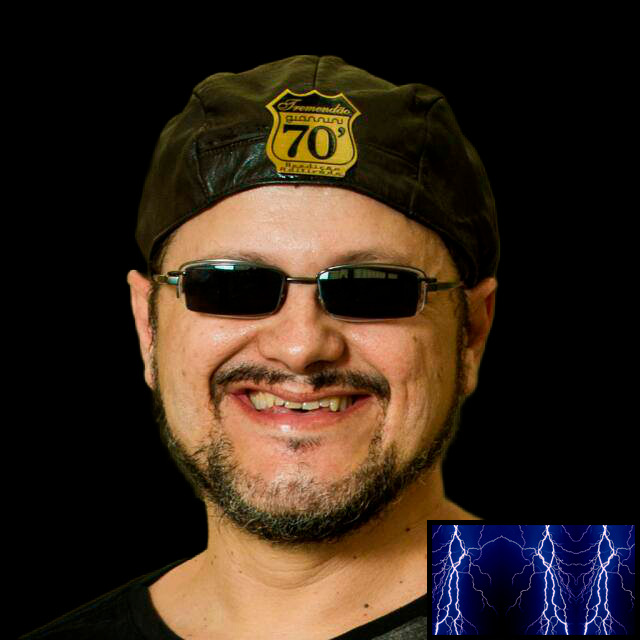} & \putimage{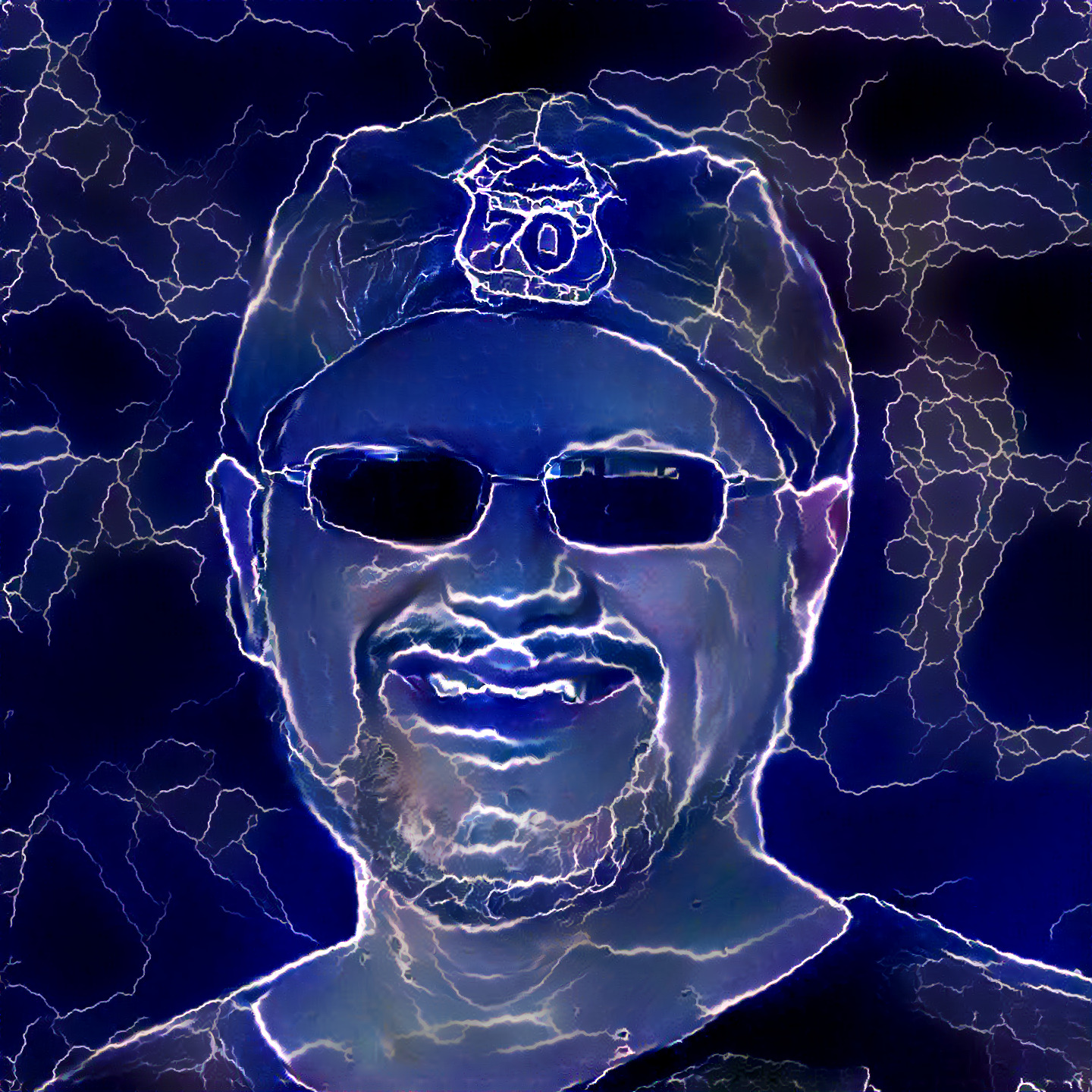} & \putimage{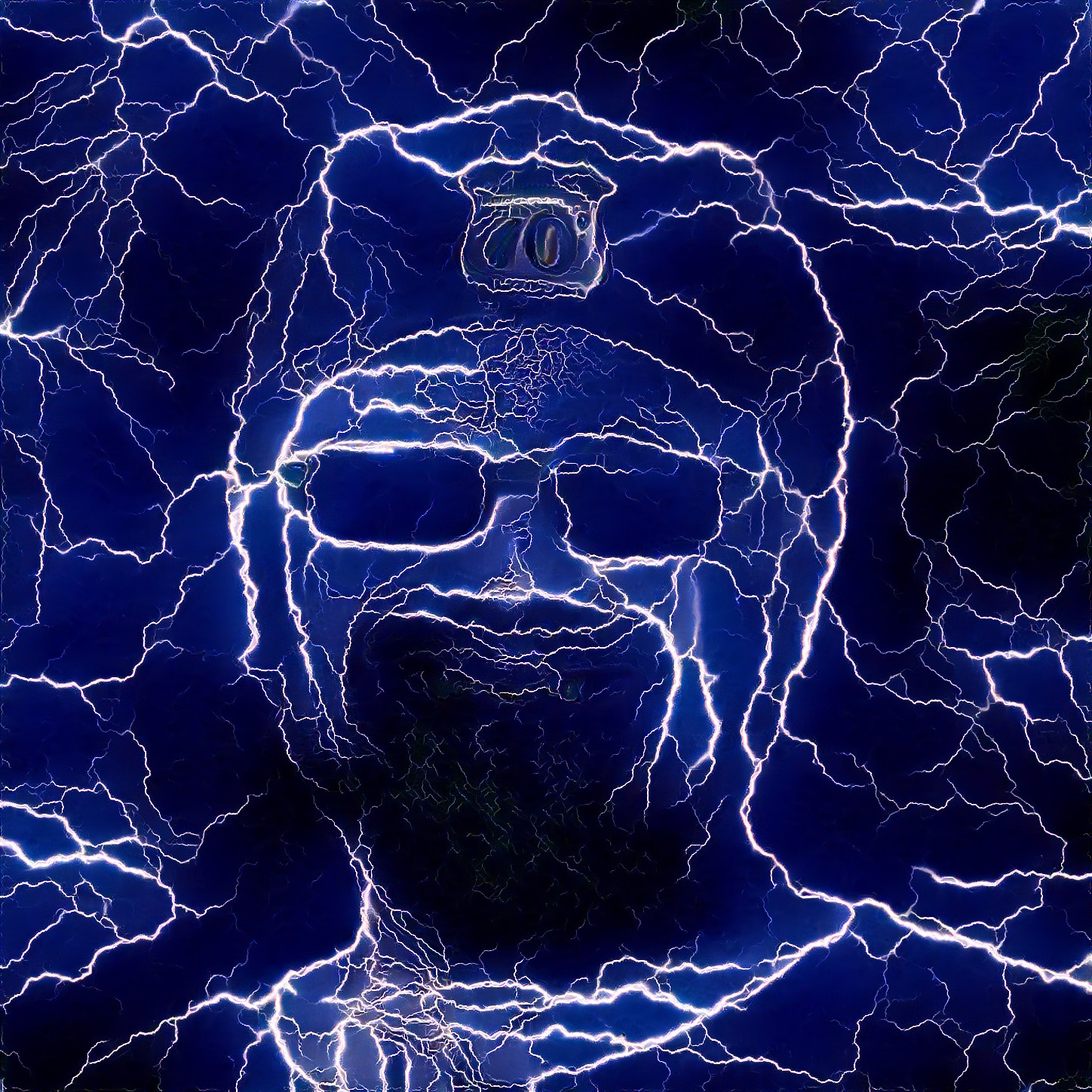} & \putimage{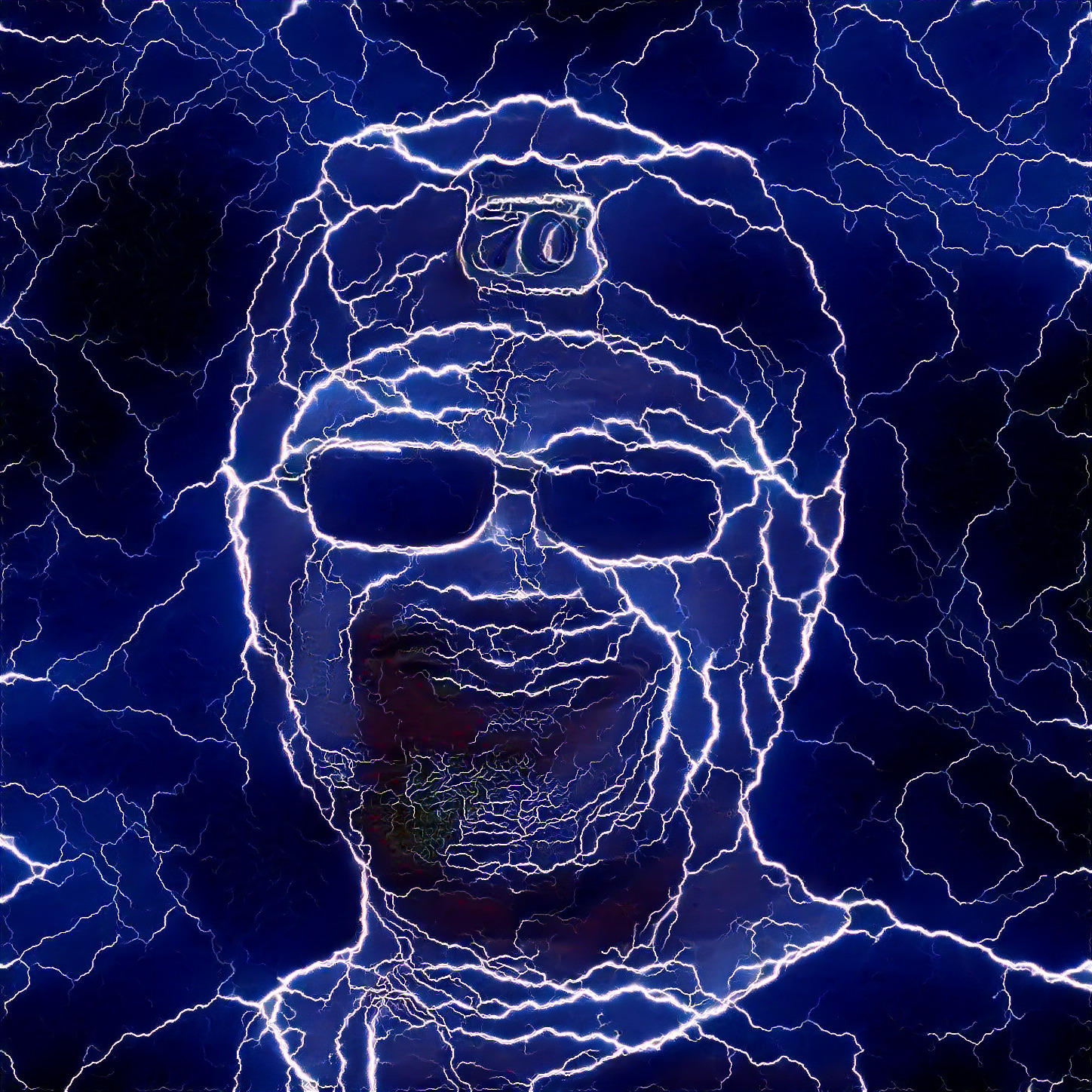} \\
            4 & \putimage{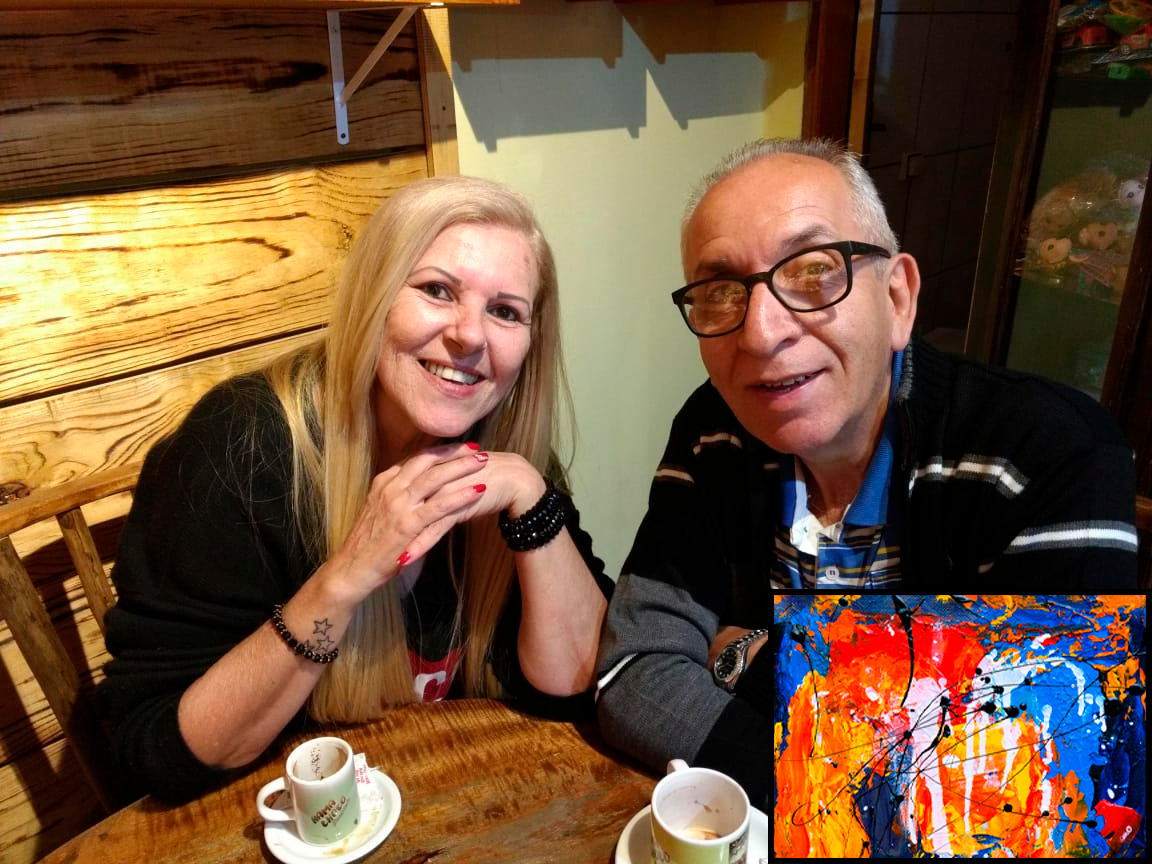} & \putimage{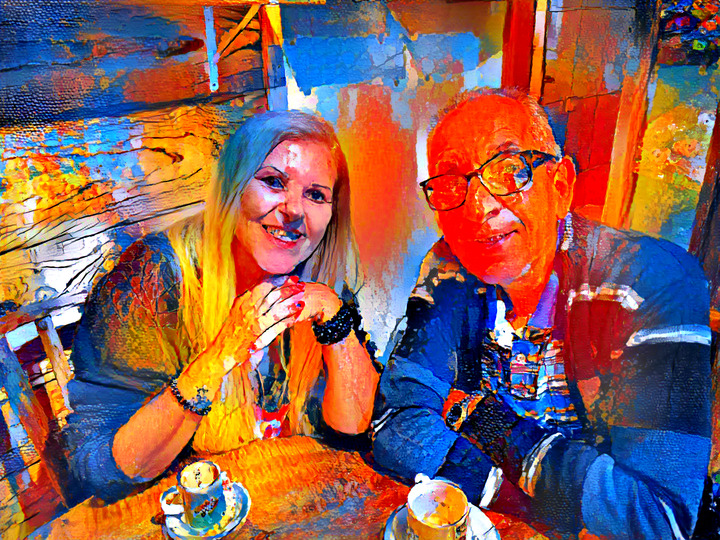} & \putimage{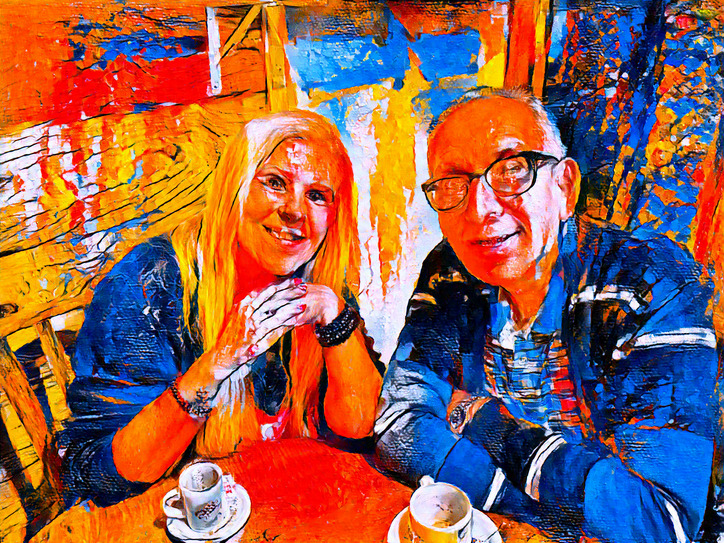} & \putimage{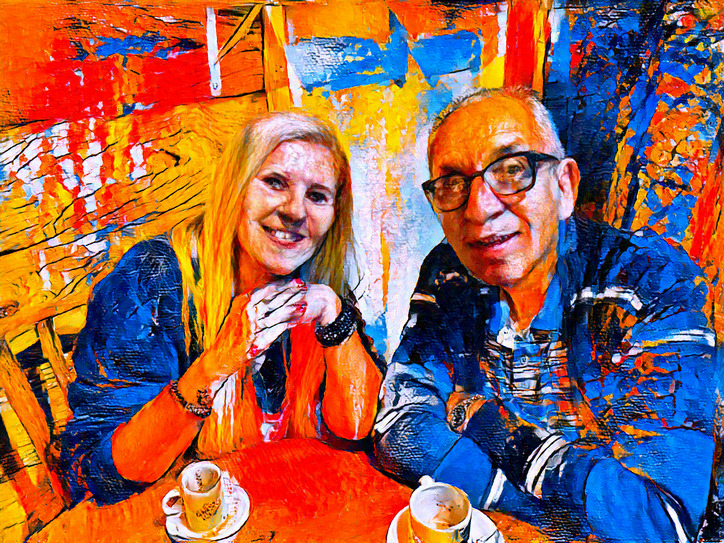} \\
            \bottomrule
        \end{tabular}

        \caption{
            Some experiments were made to compare the performance
            between the Gatys' 2016 method, the Crowson's method and the
            \emph{Portrait Stylization} method when applied to portrait
            images. The \emph{content} and \emph{style} image of each
            sample is shown in the Input column.
        }
        \label{tab:table_comparisons}
    \end{table}

\section{Conclusion} \label{sec:conclusion}

    This paper proposes improvements to the Crowson's algorithm
    to improve the quality of the results in images that contain one or more human faces.
    These improvements show that this algorithm can be optimized for specific
    image groups (e.g.\ portraits, cars, or animals) through changes in the
    \emph{total loss} function.

    The addition of auxiliary models, pre-trained
    on domain-specific tasks, can help the stylization process giving the user
    more control over how the style and content will be merged in the generated
    image.
    So, in the same way that face detectors can help in the portrait styling process,
    other domain-specific models like a pre-trained pose-estimation algorithm, could help
    in the stylization of full-body images.

    This is a great experiment to explore in future research, jointly with the possibility
    of fine-tuning the VGG-Network with a stacked dataset of open-domain and specific-domain
    images.
    This could enable a single content extraction model, optimizing the memory usage and
    the speed of the whole stylization process.

\bibliographystyle{sbc}
\bibliography{main}

\end{document}